\documentclass[runningheads]{llncs}
\usepackage[T1]{fontenc}
\usepackage{graphicx}
\usepackage{cite}
\usepackage{amsmath,amssymb,amsfonts}
\usepackage{algorithmicx}
\usepackage{graphicx}
\usepackage{textcomp}
\usepackage{float}
\usepackage[rgb]{xcolor}
\usepackage{subcaption}
\usepackage{multirow}
\usepackage{booktabs}

\usepackage{makecell}
\usepackage[colorlinks=true,allcolors=blue]{hyperref}

\usepackage{color}

\usepackage{pgfplots} 
\usepackage{afterpage} 
\usepackage{placeins}
\usepackage{nicematrix}
\usepackage{tikz}
\usetikzlibrary{shapes.geometric, arrows, fit, decorations.pathreplacing, calc, positioning}

\usepackage{pifont}
\newcommand{\xmark}{\textcolor[rgb]{0.9,0,0}{\ding{56}}}%

\usepackage[normalem]{ulem}

\usepackage{algorithm}
\usepackage[noend]{algpseudocode}
\definecolor{mutedorange}{RGB}{230, 120, 60} 
\definecolor{mutedblue}{RGB}{90, 142, 185}
\definecolor{mutedgreen}{RGB}{100, 150, 100}

\newcommand{\model}{\Phi}
\newcommand{\R}{\mathbb{R}}
\newcommand{\outputspace}{\mathcal{Y}}
\newcommand{\globalexplanation}{E}
\newcommand{\localexplainer}{L_\model}

\newcommand{\instanceset}{X}
\newcommand{\instance}{\mathbf{x}}

\newcommand{\attributionvector}{\mathbf{w}}

\newcommand{\literal}{V}
\newcommand{\attribute}{A}

\newcommand{\Language}{\mathcal{L}}
\newcommand{\setofexplanations}{\mathcal{E}}

\newcommand{\closedset}{\mathcal{C}}
\newcommand{\support}{\sigma}

\newcommand{\terms}{\mathcal{T}}
\newcommand{\dnf}{\text{DNF}}
\newcommand{\classrule}[2]{#1 \to #2}
\newcommand{\algname}{{\sc CFIRE}}
\newcommand{\longalgname}{{\sc Closed Frequent Itemset Rules from Explanations}}

\newcommand{\transactiondatabase}{\mathcal{D}}
\newcommand{\anchors}{{\sc Anchors}}
\newcommand{\anchorsk}{\anchors}
\newcommand{\anchorsinf}{\anchors-$\infty$}
\newcommand{\CEGA}{{\sc CEGA}}
\newcommand{\importantfeatures}{I}
\newcommand{\ie}{i.e.,\ }
\newcommand{\eg}{e.g.,\ }
\newcommand{\frequencythreshold}{\tau}
\newcommand{\closedfrequent}{\mathcal{C}}
\renewcommand{\indent}{\hspace*{2em}}
\newcommand{\tablespace}{0em}
\usepackage{arydshln} 

\newcommand{\hb}{2mm} 
\newcommand{\vb}{9mm} 
\newcommand{\mb}{2mm} 

\renewcommand{\xmark}{Compl.}

\makeatletter
\renewcommand{\@fnsymbol}[1]{\relax}
\makeatother
\begin{document}

\title{CFIRE: A General Method for Combining Local Explanations \thanks{Accepted at \textit{3rd World Conference on eXplainable Artificial Intelligence}}}

\author{Sebastian M\"uller (\email{semueller@uni-bonn.de})\inst{1,2}\orcidID{0000-0002-0778-9695} \and \\
Vanessa Toborek\inst{1,2}\orcidID{0009-0009-8372-8251} \and \\
Tam\'as Horv\'ath\inst{1,2,3}\orcidID{0000-0001-6852-6939} \and \\
Christian Bauckhage \inst{1,2,3}\orcidID{0000-0001-6615-2128} }

\authorrunning{S. M\"uller et al.}
\institute{University of Bonn, Bonn, Germany \and
Lamarr Institute for Machine Learning and Artificial Intelligence, Germany \and
Fraunhofer IAIS, Sankt Augustin, Germany}

\maketitle       

\begin{abstract}
We propose a novel eXplainable AI algorithm to compute faithful, easy-to-understand, and complete global decision rules from local explanations for tabular data
by combining XAI  methods with closed frequent itemset mining. 
Our method can be used with any local explainer that indicates which dimensions are important for a given sample for a given black-box decision.
This property allows our algorithm to choose among different local explainers, addressing the disagreement problem, \ie the observation that no single explanation method consistently outperforms others across models and datasets.
Unlike usual experimental methodology, our evaluation also accounts for the Rashomon effect in model explainability. To this end, we demonstrate the robustness of our approach in finding suitable rules for nearly all of the 700 black-box models we considered across 14 benchmark datasets.
The results also show that our method exhibits improved runtime, high precision and F1-score while generating compact and complete rules.

\end{abstract}

\section{Introduction}
\label{sec:introduction}

\textit{Explainable artificial intelligence} (XAI) focuses on bringing transparency to the decision-making process of black-box machine learning models (see, e.g.,\cite{molnar2022interpretableml}).
Due to their inherent interpretability (see, \eg\cite{lakkaraju2016decisionsets, letham2015stroke, wang2015falling}), \textit{decision rules} are among the most popular target languages of various XAI rule extraction methods~\cite{ribeiro2018anchors, alkhatib2022cega, guidotti2019lore}.
Several algorithms have been developed in XAI to generate \textit{local} explanations for single instances of the input space (see, \eg~\cite{lundberg2017SHAP, ribeiro2016LIME, sundararajan2017axiomatic, ribeiro2018anchors}).
While useful in some scenarios, these instance-specific local explanations fall short in terms of comprehensibility and generalisability. 
To address these shortcomings, existing efforts focus on aggregating \textit{local} explanations into \textit{global} models~\cite{ribeiro2018anchors, alkhatib2022cega}.

In this work we propose \longalgname\ (\algname), a novel local-to-global XAI algorithm for extracting rules from black-boxes trained on tabular data. It produces global rule models in disjunctive normal forms (DNFs) that are:
\newpage
\begin{itemize}
  \item[(i)] \textit{faithful},
  \item[(ii)] \textit{compact},
  \item[(iii)] \textit{complete}, \ie can explain all classes in both binary and multi-class settings,
  \item[(iv)] \textit{robust} against the Rashomon effect in XAI, and
  \item[(v)] \textit{flexible}, \ie can work with different local explainers.
\end{itemize}
While the first three properties are quite standard in the literature (see, e.g., \cite{ribeiro2018anchors, alkhatib2022cega, setzu2021glocalx, guidotti2019lore}), the last two, to the best of our knowledge, have so far been completely neglected. 
In particular, (iv) is motivated by the \textit{Rashomon effect}~\cite{breiman2001statistical} in XAI, where different yet equally well-performing models can make the \textit{same prediction} for \textit{different reasons}, with certain models even being more difficult to explain than others\cite{mueller2023rashomon}.
Property (iv) addresses this effect by requiring the method to produce equally well-performing global explanations for equally well-performing black-box models. 

Finally, property (v) is motivated by the \textit{disagreement problem}~\cite{krishna2022disagreement}, i.e. the general observation that \textit{different local explainers} produce \textit{different explanations} for the \textit{same black-box prediction}. 
Our experimental results show that the the ability of dynamically choosing the local explainer can strongly impact the quality of the resulting rule model. Hence the demand for flexibility is of crucial importance.

In brief, {\algname} works as follows: Given a black-box model, a local explainer, and a set of input samples, {\algname} first generates a set of important feature sets by using \textit{closed frequent itemset} mining~\cite{pasquier1999discovering} from the local explanations computed for the samples. Each closed frequent itemset is a subset of the feature set and thus defines a subspace of the input space. 
For each subspace, it generalizes the samples by creating a set of axis-aligned boxes, ensuring that samples with different predicted class labels are placed in separate boxes.
Finally, using the greedy heuristic for the set cover problem, it selects a compact set of boxes from this set of box systems and returns it in the form of a set of DNFs. 

A key distinction from related approaches~\cite{alkhatib2022cega} lies in the use of \textit{closed} frequent itemsets rather than \textit{all} frequent itemsets. The number of frequent itemsets grows exponentially with the number of important features, requiring some small bound on the number of features to keep computations feasible. However, as we demonstrate experimentally, this constraint leads to overly general rules. In contrast, closed frequent itemsets provide a \textit{lossless} compression of frequent itemsets and can be listed \textit{efficiently} (see, e.g., \cite{boley2010listing}) regardless of the number of important features. This allows {\algname} to generate more specific important feature combinations.

Unlike traditional surrogate models, our algorithm ensures that the resulting DNFs reflect the black-box model’s behavior in at least two key aspects: First, the properties of closed frequent itemsets ensure that the rules are directly restricted to the \textit{specific input dimensions} indicated by the local explanations.
Second, the rules provide the most specific generalizations possible: for any set of training examples inducing a box, the smallest box containing the examples is calculated. In other words, {\algname} does \textit{not} extrapolate beyond observed subspaces \textit{nor} does it compute box rules with unbounded edges.

Our experimental results on 14 benchmark datasets clearly show that {\algname} generates \textit{faithful}, \textit{compact} and \textit{complete} global rule models for both binary and multi-class problems (properties~(i), (ii) and (iii)). 
To show that {\algname} fulfills property~(iv), we evaluate it on 50 high-performing black-box models per task, confirming its robustness against the Rashomon effect.
Our results clearly show that its flexibility in choosing among multiple local explainers (property~(v)) has a large positive impact on performance in terms of properties (i)-(iii).

Finally, 
our experiments focusing on the (dis)agreement between {\algname} and the local explainers show that {\algname} is able to retain key semantic properties of the local explanations, even on unseen data.
Furthermore, compared to its strongest competitor in terms of all properties described above, we observe a runtime improvement for {\algname} by a factor of up to two orders of magnitude.

The rest of the paper is organized as follows: Section~\ref{sec:relatedwork} reviews related work, while Section~\ref{sec:preliminaries} covers the necessary background on quality metrics for rules and on closed frequent sets. In Section~\ref{sec:problemseting}, we state and motivate the problem setting. Section~\ref{sec:algorithm} details our algorithm, which is evaluated on benchmark datasets in Section~\ref{sec:experiments}. Finally, Section~\ref{sec:conclusion} concludes and outlines future research directions.
The code to reproduce all experiments is available on Github [\href{https://anonymous.4open.science/r/cfire_wcxai-018C/README.md}{anonymized repo}].

\section{Related Work}
\label{sec:relatedwork}

Building global rule-based explanations from a given black-box
model generally involves two steps: 1) \textit{rule extraction} and 2) \textit{rule composition}. Rule extraction generates candidate rules based on the behavior of the black-box model, while rule composition selects a subset of these rules to form a global explanation.

A common way to categorize explainability methods is along the axis of \textit{local} versus \textit{global} explanations. Local methods provide explanations for individual predictions, focusing on why a model made a particular decision for a single input. In contrast, global methods aim to explain the overall behavior of a model across multiple inputs or the entire input space. 

Our goal in this work is to compute a global explanation in the form of a rule-based model from local explanations. We focus on the most prominent representatives of local explanations, in particular, feature attribution methods.

Attribution methods such as KernelSHAP \cite{lundberg2017SHAP}, LIME\cite{ribeiro2016LIME}, and Integrated Gradients \cite{sundararajan2017axiomatic} assign weights to individual input features, quantifying their influence on a model's output for a given input sample. A high positive or negative value indicates that the feature strongly supports or opposes the model's prediction, respectively, while values close to zero suggest that the feature had negligible influence. Users typically focus on the most highly weighted features, but the exact magnitude of these scores is difficult to interpret \cite{krishna2022disagreement}.
To address this lack of interpretability, the authors of LIME later developed \anchors\ \cite{ribeiro2018anchors}.

{\anchors} is a local explanation technique that represents explanations in the form of local decision-rules.
Given a sample, a precision threshold, and a background distribution, \anchors\ searches for a term with the highest coverage applicable within the local neighborhood of the sample that still satisfies the precision threshold. Starting with an empty term, it generates high coverage candidate predicates to add to the term. By repeatedly drawing samples from a background distribution, it determines the most precise candidate, adds it to the final term, and starts over by generating more candidates. The process terminates once no additional candidates are found. Similarly to LIME, the background distribution can be defined over an interpretable ambient space, instead of the data space directly, and can be purely synthetic.
While the focus of {\anchors} clearly lies on the rule extraction, the authors also experimented with a simple technique to select multiple local rules to jointly explain the model behavior more globally. Given a user defined integer $k$, the method selects $k$ of the highest-precision terms where all terms cover at least one distinct sample.
We refer to the construction of a globally applicable rule model from local rules as \textit{bottom-up}.
Note that this $k$ has to be fixed in advance. This is a weakness of {\anchors} because there is no guarantee of completeness -- meaning it may not generate an explanation for every class -- if this parameter is set too low. We have observed this behavior across many model initializations.
Hence, in our experiments we will also consider an idealized variant, {\anchorsinf}, which continues to select terms until it reaches maximum coverage. This will serve as a reference for the maximal local-to-global performance of {\anchors}. While {\anchors}, in principle, can naturally handle multi-class settings, it cannot work with different local explainers (property (v)).

A very different approach to computing a rule model is taken by CEGA \cite{alkhatib2022cega}. Instead of computing rules for individual samples one by one, CEGA starts with a set of attribution scores computed for a set of input samples. Because multiple samples are considered at the same time, regardless of their position in the input space, we refer to this paradigm as \textit{top-down}. In a first step, it uses the Apriori algorithm \cite{Agrawal_etal/1997} on the attribution scores to obtain \textit{sets of input features} that are frequently marked as `important' simultaneously.
A drawback of this approach is that the cardinality of frequent itemsets considered must be limited by some small constant in advance for complexity issues. 
In the next step, it uses the training samples, corresponding labels, and sets of input features to perform association rule mining. This step determines the relationship between the actual values observed in the marked input dimensions and the class labels. This produces a set of rules that describe an association between feature values and one or multiple classes. Rules that contain more than one class label are filtered out.
The authors state that in scenarios where classes are highly imbalanced, association rule mining only finds rules for one of the classes. To counter this problem, the authors limit their method to binary problems only. This allows CEGA to use not only the positive attribution scores as evidence for one class but also the negative ones for evidence towards the \textit{other} class.
Our algorithm is not restricted to binary problems because rules are learned in a one-vs-rest scheme for each class individually.
This addresses the rule extraction from CEGA. For rule composition, CEGA distinguishes between two modes: \textit{discriminatory} and \textit{characteristic}. In the discriminatory mode, the class label is in the consequent; in the characteristic mode, it appears in the antecedent. Rules of the opposite form are omitted based on the selected mode, and the remaining rules are filtered by a confidence threshold. We note that CEGA's itemset mining based approach allows it to not only work with attribution scores as inputs but with any input that provides some form of description of input samples, including rules. In summary, {\CEGA} fulfills property (v) but not property (iii).

\section{Preliminaries}
\label{sec:preliminaries}

In this section we collect the necessary background for our work.
In particular, we discuss established performance measures for global rule-based explanations (Section~\ref{sec:prelimeval}) and provide the most important notions concerning closed frequent itemsets (Section~\ref{sec:cfi}).

\subsection{Evaluation Criteria and Methods}
\label{sec:prelimeval}
Evaluating the quality of explanations is an active research field of XAI~\cite{nauta2023co12,agarwal2022openxai}.
A fundamental contention arises between the need of explanations to be \textit{faithful} --- reflecting the actual decision process of the black-box model as best as possible --- and the need to be \textit{understandable} by the target user~\cite{jacovi2020towards, beckh2022hackfleisch}.
In Section~\ref{sec:introduction} we have formulated five natural properties that should be fulfilled by every local-to-global XAI rule extraction algorithm.
In this section we specify how we assess the performance of {\algname} in terms of these requirements.

Property~(i) from Section~\ref{sec:introduction} requires the output of {\algname} to be \textit{faithful}.
Accuracy, precision, F1-score, and coverage are established measures to assess this property in rule based explanations~\cite{ribeiro2016LIME, guidotti2019lore, alkhatib2022cega}. 
In particular, accuracy and precision address the aspect that a valid explanation needs to agree with the black-box output and coverage quantifies the extent to which a global rule-based model is even applicable in the black-box's input domain. 
We therefore assess faithfulness by reporting \textit{precision} as the accuracy of a rule at face value and \textit{F1-score} as a joint measure for error rate and coverage.

Property~(ii) requires the rules to be \textit{understandable} to the target user. 
For each class, {\algname} returns a rule in the form of a DNF with literals defined by interval constraints.
Accordingly, the comprehensibility of a global model formed by the union of such DNFs depends strongly on its size.
In case of rule-based models, this is often measured by the number of independent rules that comprise this model~\cite{alkhatib2022cega, setzu2021glocalx}. 
We follow this approach and define the \textit{size} of this kind of global rule models by the total number of terms in it, across all classes.

According to (iii), the output is expected to be \textit{complete} in the sense that all instances of the domain are covered by at least one ``powerful'' rule.
This measure is of particular importance for algorithms, including {\algname}, that can handle multi-class problems.
We quantify completeness by the proportion of the cases in which 1) the algorithm provides a rule model with above chance precision and 2) at least one rule for each of the classes.

Property (iv) requires the algorithm to show \textit{robustness} against the Rashomon effect~{\cite{breiman2001statistical}} in XAI. 
This requirement is motivated by the recent experimental study~\cite{mueller2023rashomon}, which shows that explanations obtained by the same XAI method for two different but equally well-performing models will likely diverge and that some models are easier to explain than others. The results in \cite{mueller2023rashomon} highlight that it is not sufficient to demonstrate the effectiveness of an explanation method on a single instance of a black-box hypothesis class, but that explanation methods need to be evaluated across \textit{several} instances. 
To the best of our knowledge, the robustness of the methods against the Rashomon effect has, up to now, been disregarded. 
To evaluate {\algname} for this property, we conduct all experiments for not one, but 50 equally well-performing black-box models per task and calculate mean values and standard deviations for all measures (F1-score, precision, size). \textit{Low} standard deviations in F1-scores and precision indicate robustness, while \textit{high} variability in rule size suggests the algorithm adapts to individual models.

The requirement for \textit{flexibility} (property~(v)) results from the \textit{disagreement problem}~\cite{krishna2022disagreement}.
It is the phenomenon that different local explainers may produce different explanations for the same black-box prediction, yet no single attribution method can consistently outperform others across models and datasets.
While the disagreement of feature attribution methods has been further quantified in previous work \cite{mueller2023rashomon}, its downstream effect on methods that take such attributions as input has not been studied so far. 
As a first step towards this direction, we perform an ablation study on {\algname}. We compare the results for rule models based on different local attribution methods and show that \textit{flexibility} can have a strong \textit{positive} effect on the outcome, in particular for properties (i)--(iii).

Besides the five properties discussed above, we further quantify the agreement between the rules and local explanations by measuring, for each sample, the \textit{precision} with which the applicable rules align with the dimensions identified as important by the local explanation, given that the rules assign the sample the same label as the black-box model. 

\subsection{Closed Frequent Itemsets}
\label{sec:cfi}

We recall some basic notions and results from \textit{frequent pattern mining} (see, e.g., \cite{aggarwal2014frequent}).
For a finite ground set $U$, let $\transactiondatabase$ be a multiset of subsets of $U$ (i.e., a subset of $U$ may occur in $\transactiondatabase$ with muliplicity greater than one).
Given $\transactiondatabase$ over $U$ and a (relative) frequency threshold $\tau \in (0,1]$, the problem of \textit{frequent itemset mining}~\cite{Agrawal_etal/1997} is to generate all subsets of $U$ that are subsets of at least $t$ sets in $\transactiondatabase$ for $t = \lceil \tau |\transactiondatabase|\rceil$. 
The subsets of $U$ satisfying this property are referred to as \textit{frequent (item)sets}.
It follows that if $F \subseteq U$ is frequent then all proper subsets of $F$ are also frequent, implying that the number of frequent itemsets is exponential in the cardinality of a frequent itemset of maximum cardinality.
Instead of frequent itemsets, one may consider the collection (also called family) of closed frequent itemsets~\cite{pasquier1999discovering}; a subset $F \subseteq U$ is \textit{closed frequent} if it is frequent and for all $F' \supsetneq F$, there exists $T \in \transactiondatabase$ with $F \subseteq T$ and $F' \nsubseteq T$.

As an example, consider the ground set $U = \{a,b,c,d,e\}$ and the family 
$$\transactiondatabase = \{ bce, abde, abde, abce, abcde, bcd\} \enspace ,$$
where the strings in $\transactiondatabase$ represent sets of elements (e.g., $bce$ denotes $\{b,c,e\}$).
Note that $\transactiondatabase$ is a multiset, as $abde$ appears more than once in $\transactiondatabase$.
Let the frequency threshold be $\tau =0.5$.
Then $bde$ is frequent, as it is a subset of $3= \lceil 0.5 \cdot 6\rceil$ sets in $\transactiondatabase$. 
However, $bde$ is not closed. 
Indeed, if we extend $bde$ by $a$, then $abde$ is a subset of all three sets in $\transactiondatabase$ containing $bde$. 
In contrast, $bce$ is not only frequent, but also closed.
For all proper supersets of $bce$, none is a subset of \textit{all} three sets in $\transactiondatabase$ that contain $bce$ (i.e., $bce,abce,abcde$).
One can check for $\transactiondatabase$ and $\tau$ above that while the number of frequent itemsets is 19, that of closed frequent itemsets is only 7.
In Section~\ref{sec:algorithm} we will discuss some attractive algebraic and algorithmic properties of closed frequent sets that are also utilized by our algorithm.

\section{Problem Setting}
\label{sec:problemseting}

In this section we define the problem setting considered in this work.
Given a black-box classifier model $\model$ mapping the domain $\R^d$ to a finite set of target classes for some positive integer $d$, our goal is to construct a \textit{global explanation} formed by a set of rule-based explanations, each computed for one of the target classes. 
In the following, the set $\{1,\ldots,d\}$ is denoted by $[d]$. 

\textbf{Explanation Language} 
We consider class explanation rules of the form 
$
\classrule{E}{c} \enspace ,
$
where the antecedent $E$, the \textit{explanation}, is a DNF and the consequent $c$ is one of the target classes. 
More precisely, the \textit{language} $\Language$ for the explanations is formed by the class of DNFs over variables defining \textit{interval constraints}. 
That is, all variables $\literal$ in all \textit{terms} (i.e., conjunctions) of a DNF in $\Language$ are of the form $\attribute_i \in [a,b]$ for some $a,b \in \R$ and $i \in [d]$, where $\attribute_i$ denotes the $i$-th feature. 
Without loss of generality, we require that each term of a DNF contains at most one interval constraint for $\attribute_i$, for all $i\in[d]$. 
The terms in $\globalexplanation$ represent \textit{boxes}, i.e., axis-aligned hyperrectangles, in subspaces of $\R^d$.
Thus, a rule $\classrule{E}{c}$ with $E \in \Language$ approximates the set of instances with predicted class $c$ 
by the union of a finite set of such boxes. 
Although this limitation of $\Language$ may seem too restrictive at first glance, our experimental results and those in \cite{ribeiro2018anchors} using the same language clearly show that $\Language$ has a sufficiently large expressive power.
Needless to say, DNFs in $\Language$, containing only a few terms, are undoubtedly \textit{easy} for domain experts to interpret, provided the terms do not contain too many literals (i.e., interval constraints).
We emphasize that for all classes $c$, we are interested in rules that explain the predicted and \textit{not} the (unknown) true class label.

\textbf{The Problem} 
To compute $E\in\Language$ for a rule $\classrule{E}{c}$ for some class $c$, the algorithm receives as input a finite subset $\instanceset$ of the domain generated by the \textit{unknown} target distribution.
The input set $\instanceset$ is required to contain at least one instance with predicted class $c$.
We assume that the algorithm has access to the \textit{black-box model} to be explained and to a \textit{local explainer} algorithm. 
The black-box model, denoted $\model$, is a function mapping the domain $\R^d$ to $\outputspace$, where $d$ is the dimension of the domain and $\outputspace$ is the set of target classes. 
We deal with \textit{classification} models, i.e., $\outputspace$ is some finite set.
For an input sample $\instance \in \R^d$, it returns the \textit{predicted} class $\model(\instance) \in \outputspace$ of $\instance$.
The \textit{local explainer}, denoted $\localexplainer$, returns for the black-box model $\model$ and $\instance$ an \textit{attribution vector} $\attributionvector \in \R^d$. 
It is a $d$-dimensional vector with real valued entries, where entry $i$ indicates how much feature $i$ contributes to the black-box model's prediction of the class of $\instance$. 
Finally, the set of instances of $\instanceset$ predicted as $c$ is denoted by $\instanceset_{c}$, i.e., 
$$
\instanceset_{c}=\{\instance \in \instanceset: \model(\instance) = c\}
\enspace .
$$
Note that $\instanceset_c$ depends on $\model$. We omit $\model$ from the notation because it is always clear from the context.
Our goal is to compute a \textit{compact} explanation for each class $c$ that is \textit{consistent} with the predicted labels of all instances in $\instanceset$, i.e., it covers all instances in $\instanceset_c$ and none of the elements of $\instanceset \setminus \instanceset_c$.
More precisely, we consider the following problem:
\begin{problem}
\label{problem:main}
\textit{Given} a black-box model $\model: \R^d \to \outputspace$ for some positive integer $d$ and finite set $\outputspace$ of class labels, a local explainer $\localexplainer$ for $\model$, and a finite set $\instanceset \subset \R^d$ with $\instanceset_{c} \neq \emptyset$ for all $c \in \outputspace$, \textit{compute} a set 
$$
\setofexplanations = \{\classrule{\globalexplanation_c}{c} \text{ with } \globalexplanation_c \in \Language\}_{c \in \outputspace}
$$ 
of class explanations that is 
\begin{itemize}
  \item[(i)] consistent with $\model$ on $\instanceset$, i.e., for all $\instance \in \instanceset$, $c \in \outputspace$, and $\classrule{E}{c} \in \setofexplanations$, $\instance$ satisfies $\globalexplanation$ iff $\model(\instance) = c$ and
  \item[(ii)] all explanations in $\setofexplanations$ have the smallest size with respect to this property, i.e., for all $c \in \outputspace$, there is no $\classrule{\globalexplanation'}{c}$ with $\globalexplanation' \in \Language$ that satisfies the consistency condition above and $\globalexplanation'$ has strictly less terms than $\globalexplanation$. 
\end{itemize}
\end{problem}

We note that Problem~\ref{problem:main} is NP-complete. 
Indeed, it is in NP and the NP-hardness follows from a straightforward reduction from the following decision problem from computational learning theory: \textit{Given} disjoint sets $P,N \subseteq \{0,1\}^d$ for some $d$ and a positive integer $k$, \textit{decide} whether there exists a $k$-term-DNF (i.e., which consists of $k$ terms) that is consistent with $P$ and $N$ (i.e., which is satisfied by all instances in $P$ and by none in $N$). This consistency problem is NP-complete already for $k=2$~\cite{pitt1988computational}.

The negative complexity result above implies that we need to resort to some \textit{heuristic} approach to approximately solve Problem~\ref{problem:main}. 
In the next section we present such a heuristic algorithm.
Among other things, it makes use of some nice algebraic and algorithmic properties of \textit{closed sets} that are relevant for our context. 
Note that Problem~\ref{problem:main} allows the rules in $\setofexplanations$ to be approximated for each class separately, as we only require the explanations in the rules to be consistent with the predicted classes of the instances in $X$; the family of subsets of $\R^d$ represented by the class explanations in $\setofexplanations$ are \textit{not} required to be pairwise disjoint.
This observation allows the instances in $X$ with the same predicted class to be handled separately for every class.

\newpage
\section{The Algorithm}
\label{sec:algorithm}

\begin{figure*}[t]
  \centering
  \label{fig:overview}
  \resizebox{\textwidth}{!}{
    \tikzstyle{artifact} = [ellipse, minimum height = 15mm, text width=30mm, text centered, line width = 0.5mm, draw={rgb,255:red,78; green,165; blue,205}, fill={rgb,255:red,78; green,165; blue,205}, fill opacity=0.3, text opacity=1]

\tikzstyle{method} = [rectangle, draw, text centered, draw={rgb,255:red,67; green,175; blue,105}, line width = 0.5mm, fill={rgb,255:red,67; green,175; blue,105}, fill opacity=0.3, text opacity=1, minimum height = 15mm, rounded corners, text width=50mm]

\newcommand{\arrowwidth}[0]{1.5pt}
\tikzstyle{arrow} = [thick, ->, >=stealth, line width=\arrowwidth]

\begin{tikzpicture}[node distance=20mm]

        \node(data)[artifact, text width=15mm]{input set ($\instanceset \subsetneq \R^d$)};
        \node(blackbox)[method, text width=15mm, right = 5mm of data]{trained black-box model ($\model$)};
        \node(explain)[method, text width=15mm, left = 5mm of data]{local explainer ($\localexplainer$)};

        \node[method, text width=15mm, below=10mm of data] (attributes) {get important features};
        \node[method, text width = 25mm, right=10mm of attributes] (closed) {enumerate important closed frequent featuresets};
        \node[method, text width = 20mm, right=10mm of closed] (rules) {learn partial class rules};
        \node[method, text width = 25mm, right=10mm of rules] (select) {select and compose partial class rules};
        \node[artifact, text width = 15mm, right=17mm of select] (dnf) {class DNFs};
        \node[artifact, text width = 15mm, below=10mm of dnf] (global) {global rule model ($\setofexplanations$)};

        \node[inner ysep=3mm, inner xsep=3mm, fit=(attributes)(closed)(rules)(select)(dnf), draw={rgb,255:red,239; green,105; blue,219}, line width = 0.5mm, rounded corners, text height=20mm, align=center] (us) {};

        \draw[arrow] (data) -- (attributes);
        \draw[arrow] (blackbox) -- (attributes);
        \draw[arrow] (explain) -- (attributes);
        \draw[arrow] (attributes) -- (closed);
        \draw[arrow] (closed) -- (rules);
        \draw[arrow] (rules) -- (select);
        \draw[arrow] (select) -- node[midway, above]{collect all} (dnf);
        \draw[arrow] (dnf) -- (global);

        \coordinate (intermediate1) at ($(select.south)+(0, -12mm)$);
        \coordinate (intermediate2) at ($(attributes.south)+(0, -12mm)$);

        \draw[-, thick, line width = \arrowwidth] (select.south) -- (intermediate1);
        \draw[-, thick, line width = \arrowwidth] (intermediate1) -- node[midway, below]{for each class} (intermediate2);
        \draw[arrow] (intermediate2) -- (attributes.south);

        \end{tikzpicture}
  }
  \caption{Schematic overview of our algorithm {\algname}. Given black-box model $\model$, local explainer $\localexplainer$, and input data $\instanceset$, our algorithm computes DNFs as explanations for each class independently.}
\vspace{-1em}
\end{figure*}
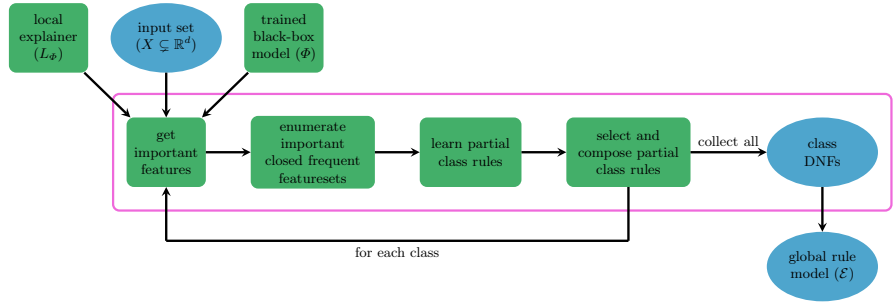

In this section we present our \textit{heuristic} algorithm {\longalgname} (\algname) for Problem~\ref{problem:main}. 
A schematic representation and the pseudocode of the algorithm are given in Figure~\hyperref[fig:overview]{1} 
and Alg.~\ref{alg:main}, respectively.\footnote{For simplicity we omit the hyperparameters from the description because they were constant in all our experiments (see Section~\ref{sec:experiments})}
The input to the algorithm consists of a black-box model $\model$, a local explainer $\localexplainer$, and a finite subset $X$ of $\R^d$, as defined in Problem~\ref{problem:main}.
It is important to emphasize that the algorithm can be parameterized by \textit{any} local explainer that computes explanations which specify the dimensions that are important for a given black-box prediction (cf. property (v) on flexibility in Section~\ref{sec:introduction}). In this work we will consider attribution methods for their popularity and ease of use.
The downstream effect of the \textit{disagreement problem}, that is, the impact of using \textit{different} local explainers, is evaluated experimentally and discussed in Section~\ref{sec:disagreement}.
The algorithm processes every target class $c \in \outputspace$ separately (cf. lines~\ref{line:startouterloop}--\ref{line:endouterloop}) and returns a set of class explanations, one for each class, in $\setofexplanations$ (cf. line~\ref{line:output}).
The explanation for the current class $c$, variable $\dnf$ in Line~\ref{line:declareDNF}, is computed in five main steps, as described below.

\textbf{\textbf{(Step~1) Initialization}} 
We initialize $\terms$, the set of candidate terms for the explanation of class $c$, and 
$\transactiondatabase$, a multiset of subsets of the feature set, by the empty set.
Then, we compute $\instanceset_c$, which contains all input samples from $\instanceset$ that were predicted as class $c$ by the black-box model $\model$ (cf. lines~\ref{line:init}--\ref{line:classblock}).

\textbf{\textbf{(Step~2) Selection of Important Features}} In this step we calculate for all instances $\instance \in \instanceset_c$ the set of features that are important for the class prediction of $\instance$.
Formally (cf. lines~\ref{line:startloopimportantattributes}--\ref{line:attributeselection}), for all $\instance \in \instanceset_c$, the algorithm calls the local explainer $\localexplainer$ for $\instance$. 
By condition, it returns an attribution vector $\attributionvector \in \R^d$ (i.e., $\attributionvector = \localexplainer(\instance)$).
Normalizing the importance scores in $\attributionvector$ by the sum of their absolute values, a feature is regarded as \textit{important} if its absolute value is greater than a threshold $\iota$ after normalization. As a rule of thumb, we used $\iota = 0.01$ in \textit{all} of the experiments.

\begin{algorithm}[t]
\caption{\algname}
\label{alg:main}
\textbf{Input:} black-box model $\model: R^d \to \outputspace$ for some positive integer $d$ and finite set $\outputspace$, local explainer $\localexplainer$ for $\model$, finite set $\instanceset \subset \R^d$ with $\instanceset_c \neq \emptyset$ for all $c \in \outputspace$\\
\textbf{Output}: $\setofexplanations = \{\dnf_c \to c :c \in \outputspace\}$
\begin{algorithmic}[1]
  \State $\setofexplanations \gets \emptyset$
  \ForAll{$c \in \outputspace$}
    \label{line:startouterloop}
    \State $\terms, \transactiondatabase \gets \emptyset$ 
    \Comment{$\transactiondatabase$ is a multiset}
    \label{line:init}
    \State $\instanceset_c \gets \{\instance \in \instanceset : \model(\instance) = c\}$ 
    \label{line:classblock}
    \ForAll{$\instance \in \instanceset_c$}
      \label{line:startloopimportantattributes}
      \State $T_\instance \gets 
        \Call{Get\_Important\_Attributes}{\localexplainer(\instance)}$
      \label{line:attributeselection}
      \State add $T_\instance$ to $\transactiondatabase$
      \label{line:endloopimportantattributes}
    \EndFor
    \State $\closedset \gets \Call{Enumerate\_Closed\_Frequent\_Sets}
    {\transactiondatabase}$
    \label{line:closeditemsetmining}
    \ForAll{$F \in \closedset$}
      \State $\terms \gets \terms \cup \Call{Learn\_Rules}{F,\support(F),\instanceset\setminus\instanceset_c}$
    \EndFor
    \State $\dnf \gets \Call{SelectAndComposeRules}{\terms,\instanceset_c}$
    \label{line:declareDNF}
    \State add $\dnf\to c$ to $\setofexplanations$
  \EndFor
  \label{line:endouterloop}
  \State \Return $\setofexplanations$
  \label{line:output}
\end{algorithmic}
\end{algorithm}

\textbf{\textbf{(Step~3) Closed Frequent Feature Sets}} For each target class, only features that are \textit{important} for the prediction of \textit{several} instances in the block of the class are considered as candidates for the interval constraints in the class explanations.
The frequency constraint is necessary for the explanations' \textit{compactness}, as each selected feature will induce at least one interval constraint.  

Accordingly, our goal is to generate a family of feature sets in which \textit{all} features in a set contribute significantly to the prediction of \textit{several} instances in the block.
In other words, we aim to compute subsets of features, each being a candidate for inducing a term in the DNF, where the features within a particular subset \textit{jointly} contribute significantly to the class prediction for \textit{several} instances.
Note that this family can be incomplete in the sense that the binarized attribution vector of an instance in $X$ is not covered by any of these frequent feature sets. 
However, this does not imply that none of the interval constraints computed from the frequent sets will cover the instance itself.
When regarding the binarized attribution vectors as the characteristic vectors of subsets of the entire feature set, the above problem of generating all frequent sets of important features is precisely an instance of the \textit{frequent itemset mining} problem (see, also, Section~\ref{sec:cfi}): Given a transaction database, where each transaction is a subset of the set of all items and a relative frequency threshold, the task is to generate the set of all frequent itemsets, i.e., the subsets that appear in at least a certain number of the transactions specified by the threshold.

As already mentioned, each frequent set of important features induces a term in the class explanation DNF by taking the conjunction of the interval constraints computed for the features in the set.
Although this idea may seem straightforward at first glance, it raises a serious complexity problem: If there exists a frequent set of cardinality $k$, then the number of frequent itemsets is $\Omega(2^k)$, as all subsets of a frequent set are also frequent. 
To overcome this problem, one may control the number of frequent itemsets by bounding the size of the frequent itemsets by some small constant $s$.
However, this restriction results in terms that contain at most $s$ interval constraints. 
Hence, the corresponding boxes of $\R^d$ are unbounded in at least $d-s$ coordinates (features). 
For example, in the open source code for CEGA~\cite{alkhatib2022cega}, which we use in our experiments, the number of literals in a term is bounded by 3 (cf. Section~\ref{sec:relatedwork}).
Our experimental results presented in Section~\ref{sec:experiments} clearly show that the corresponding boxes are too general, implying that CEGA is unable to generate global class explanations of high accuracy.

We tackle this complexity issue by considering \textit{closed} frequent itemsets. 
Recall from Section~\ref{sec:cfi} that a frequent itemset is closed if it is maximal for the following property: its extension by a new item leads to a drop in the number of supporting transactions. 
Closed frequent itemsets have several advantages over frequent itemsets that can be utilized in our application context:
\begin{itemize}
\vspace{-0.em}
\item[(a)] For any frequent itemset $F$ with a support set (i.e., the set of instances in $\instanceset_c$ with important feature sets containing $F$), there exists a \textit{unique} closed itemset with the same support set as $F$. Thus, closed frequent itemsets are a lossless representation of frequent itemsets.
\item[(b)]
The above mentioned maximality property of closed frequent itemsets implies that the boxes determined by the interval constraints for the features in closed frequent sets are more specific (i.e., have a lower degree of freedom) than those in frequent itemsets, addressing the problem of overly general boxes mentioned above.
\item[(c)] The number of closed frequent itemsets can be exponentially smaller than that of frequent itemsets.
\item[(d)] The set of closed frequent itemsets can be generated \textit{efficiently} (more precisely, with polynomial delay, meaning that there is an algorithm such that the delay between two consecutive closed frequent itemsets generated by the algorithm is bounded by a polynomial of the size of the input).
\end{itemize}
Properties (a), (c), and (d) together imply that closed frequent itemsets form a \textit{lossless} representation of frequent itemsets, can achieve an exponentially large compression ratio, and can be listed efficiently, independently of their maximum cardinality (see, e.g., \cite{boley2010listing,gely2005generic,pasquier1999discovering}). 
These nice algebraic and algorithmic properties together with (b) make closed frequent itemsets more suitable for our purpose than ordinary frequent itemsets.

\textit{Formally}, for $c \in \outputspace$ and instance $\instance \in \R^d$, let $\importantfeatures(\instance)$ be the set of important features selected in Step~2 for $\instance$ and denote 
$$\transactiondatabase_c = \{\importantfeatures(\instance) : \instance \in \instanceset_c\}$$
the family of important feature sets for the instances in $\instanceset_c$.
The \textit{support} of a set $F$ of features in $\transactiondatabase_c$ is the family of sets in $\transactiondatabase_c$ that contain $F$, i.e., 
$$
\support(F) = \{ S \in \transactiondatabase_c : F \subseteq S \}
\enspace .
$$
We note that $\transactiondatabase_c$ and $\support(F)$ are regarded as multisets.
For $\transactiondatabase_c$ and relative frequency threshold $\frequencythreshold \in (0,1]$, let $\closedfrequent_c$ denote the family of closed frequent sets of $\transactiondatabase_c$ for $\frequencythreshold$, i.e., for all subsets $F,F'$ of the set of features with $F \subsetneq F'$,
$$
F \in \closedfrequent_c \iff \frac{|\support(F)|}{|\transactiondatabase_c|} \geq \frequencythreshold \text{ and } \support(F') \subsetneq \support(F) \enspace .
$$
It holds that $\closedfrequent_c$ can be computed in time polynomial in the combined size of $\transactiondatabase_c$ and the cardinality of the set of all features (see, e.g., \cite{boley2010listing}).
In the realization of 
line ~\ref{line:closeditemsetmining} we use the closed frequent set generation algorithm in \cite{gely2005generic}.
It is a highly effective enumeration algorithm based on the divide and conquer paradigm.
This paradigm implies some nice algorithmic properties that are utilized by our implementation of Alg.~\ref{alg:main}. For \textit{all} experiments, $\frequencythreshold = 0.01$ was used.

\textbf{\textbf{(Step~4) Learning Incomplete Class Explanations}} 
\label{sec:partialrules}
For a closed frequent feature set $F$ generated for class $c$, we take the projection of all training examples to the subspace in {$\R^d$} spanned by the features in $F$.
Let $X'$ denote the set of training examples $\instance$ with predicted class $c$ such that the important feature set $\importantfeatures(\instance)$ of $\instance$ supports (i.e., contains) $F$, that is,
$$
X' = \{ \instance \in \instanceset_c | F \subseteq \importantfeatures(\instance) \} \enspace .
$$
Finally, let $B$ be the smallest box in the subspace induced by $F$ that contains the projections of the elements in $X'$. 
If $B$ is consistent with the training examples predicted as classes different from $c$, 
we associate $F$ with the term representing this box.
Otherwise, we approach the problem as a concept learning task by applying a decision tree learning algorithm.
The goal of the decision tree is to refine $B$ with subboxes such that the projections of all instances in $X'$ is still contained by at least one subbox and the union of the subboxes is consistent with the examples in $X \setminus X_c$. 
For this case we associate $F$ with the disjunction of the terms representing the subboxes. 
We allow a certain amount of inconsistency to avoid too complex decision trees and hence overfitting rule models.

\textbf{\textbf{(Step~5) Learning Complete Class Explanations}} 
The set of terms computed in Step~4 for a given class $c$ represents a set of boxes in $\R^d$, each (almost) consistent with $X \setminus X_c$. 
Each term is associated with the set of samples in $\instanceset_c$ that satisfy it (or equivalently, whose corresponding projections are contained by the corresponding box). By construction, a particular instance in $X_c$ can satisfy several terms. 
The goal is to select a smallest set of terms such that the union of their associated samples covers all instances in $\instanceset_c$. 
Note that this selection problem can be regarded as an instance of the \textit{set cover} problem.
Since this problem is NP-hard, we resort to the standard greedy heuristic designed for it (see, e.g., \cite{cormen2022introduction}). 
Our results in Section~\ref{sec:disagreement} show that the selected rules reproduce the characteristics of the local explanations surprisingly well.
Finally, we return the explanation for class $c$ by taking the disjunction of the set of terms selected with this heuristic.

\textbf{Using {\algname} with Multiple Local Explainers} Given multiple $\localexplainer$, we compute a set $\setofexplanations_{\localexplainer}$ of class explanations for each of them and select the one, denoted $\setofexplanations$, that achieves the highest accuracy on $\instanceset$.

\textbf{Applying $\setofexplanations$}  For an instance $\instance$, we predict its class by the rule $\globalexplanation_c \to c \in \setofexplanations$ that is satisfied by $\instance$. In case that more than one $\globalexplanation_c$ is satisfied by $\instance$, we break the tie by choosing the rule where the term satisfied by $\instance$ has the highest accuracy.

\section{Experimental Evaluation}
\label{sec:experiments}

\begin{table}[!t]

  \caption{Overview of datasets used in the experiments. $|\outputspace|$ describes the number of classes, $d$ the dimensionality of the data, and $|\instanceset|$ the cardinality of the input set. $\model$ reports the average black-box accuracies on $\instanceset$.}
  
  \centering
  \label{tab:datasets_combined}
  \footnotesize
  
  \begin{tabular}{cc}
  \vspace{\tablespace}
    \begin{subtable}[t]{0.48\linewidth}
      \centering
      \label{tab:datasets_binary}
      \vspace{\tablespace}
      \caption{Binary classification datasets.}
      \resizebox{\linewidth}{!}{%
      \begin{tabular}{l c c c} \toprule
       Dataset & $d$ & $|\instanceset|$ & $\model$ \\ \midrule
       \cite{btsc_176} btsc & 4 & 150 & 0.81 \tiny{$\pm$0.00} \\ 
       \cite{breastw} breastw & 9 & 1000 & 0.99 \tiny{$\pm$0.00} \\ 
       \cite{agarwal2022openxai} heloc & 23 & 1975 & 0.73 \tiny{$\pm$0.00} \\ 
       \cite{steel_plates_faults_198_spf} spf & 24 & 389 & 0.76 \tiny{$\pm$0.01} \\ 
       \cite{breast_cancer_wisconsin_(diagnostic)_17} breastcancer & 30 & 114 & 1.00 \tiny{$\pm$0.00} \\ 
       \cite{ionosphere_52} ionosphere & 34 & 71 & 0.96 \tiny{$\pm$0.01} \\ 
       \cite{spambase_94} spambase & 57 & 921 & 0.94 \tiny{$\pm$0.00} \\ 
       \bottomrule
      \end{tabular}
      }
    \end{subtable} &
    \begin{subtable}[t]{0.48\linewidth}
      \centering
      \label{tab:datasets_multiclass}
      \vspace{\tablespace}
      \caption{Multiclass classification datasets.}
      \resizebox{\linewidth}{!}{%
      \begin{tabular}{l c c c c} \toprule
       Dataset & $|\outputspace|$ & $d$ & $|\instanceset|$ & $\model$ \\ \midrule
       \cite{iris_53} iris & 3 & 4 & 38 & 0.92 \tiny{$\pm$0.01} \\ 
       \cite{autouniv} autouniv & 3 & 5 & 140 & 0.42 \tiny{$\pm$0.01} \\ 
       \cite{abalone_1} abalone & 3 & 7 & 627 & 0.64 \tiny{$\pm$0.00} \\ 
       \cite{wine_109} wine & 3 & 13 & 36 & 1.00 \tiny{$\pm$0.00} \\ 
       \cite{statlog_(vehicle_silhouettes)_149} vehicle & 4 & 18 & 170 & 0.81 \tiny{$\pm$0.01} \\ 
       \cite{dry_bean_602} beans & 7 & 16 & 1050 & 0.91 \tiny{$\pm$0.00} \\ 
       \cite{diggle1990time} diggle & 9 & 8 & 124 & 0.96 \tiny{$\pm$0.00} \\ 
       \bottomrule
      \end{tabular}
      }
    \end{subtable}
  \end{tabular}
  \vspace{-1em}
\end{table}

In this section we present our experimental results. 
In particular, we compare the performance of {\algname} against that of \anchors~\cite{ribeiro2018anchors} and \CEGA~\cite{alkhatib2022cega}, and show that our algorithm fulfills all five properties required in Section~\ref{sec:introduction}.

\textbf{Datasets and Black-Box Models} 
We consider seven binary and seven multi-class learning tasks.
Table \ref{tab:datasets_combined} provides a summary of the datasets and performance of our black-box NNs. 
The corresponding datasets are publicly available at either \href{https://www.openml.org/}{OpenML.org}, \href{https://archive.ics.uci.edu/}{UCI Machine Learning Repository} or \href{https://doi.org/10.7910/DVN/HIBCFL}{Harvard Dataverse}. 
In some cases categorical variables trivialised the learning problem; these features were removed from the respective datasets. All datasets obtained in this way became subsets of $\R^d$ for some $d$.
Each dataset was split into three parts:
Roughly 80\% were used to train the black-box neural networks (NNs), 10\% were used as input to the algorithms (i.e., $\instanceset$), and the remainder for testing. 
For very small datasets (such as iris or wine), the split was 60-20-20.
We train 50 NNs for each task, all of which achieve similarly good performance (see Table~\ref{tab:datasets_combined}). 
Since {\CEGA} is not applicable to multi-class tasks, we compare our method {\algname} against the {\anchors} variants only in those cases.
All algorithms were given the same input set $\instanceset$ to compute global rule models. 

\textbf{Setup and Hyperparameters} For \textit{all} experiments with {\algname}, we set the importance threshold $\iota = 0.01$, the frequency threshold $\tau = 0.01$, and the maximum depth of the decision trees learned in Step~4 (cf. Section~\ref{sec:partialrules}) to $7$. 
As motivated by the disagreement problem, we run {\algname} with three local explainers: KernelSHAP (KS)~\cite{lundberg2017SHAP}, LIME (LI)~\cite{ribeiro2016LIME}, and Integrated Gradients (IG)~\cite{sundararajan2017axiomatic}. For KS, LI we use the implementation provided in \href{https://github.com/pytorch/captum}{Captum}. For each black-box we choose the {\algname} output for a single $\localexplainer$ as described in the previous section.

The results for {\anchors} were computed using the \href{https://pypi.org/project/anchors/}{anchors implementation} available on PiPy. The precision threshold was set to 0.9 for all experiments.
To obtain a global model from {\anchors},  we use the greedy selection approach outlined in \cite{ribeiro2018anchors}. It requires the user to set an upper bound $k$ on the number of rules to be selected \textit{in advance}. 
For a fair comparison at the same \textit{compactness-level}, we set $k$ to the same number of terms that {\algname} computed. 
To compare against the best \textit{theoretical} result for {\anchors}, we also report results with an unconstrained $k$, denoted by {\anchorsinf}. This \textit{idealized} version can select as many terms computed on $\instanceset$ as needed to reach maximum coverage on the input set. Since {\anchors} does not provide a prediction function, we use the same strategy as for {\algname} for both variants of {\anchors}.

To compute the CEGA explanations, we used the code available \href{https://github.com/amrmalkhatib/CEGA?tab=readme-ov-file}{on Github}. CEGA was also run with all three local explainers. 
Here, the maximum length parameter (i.e., the upper bound on the cardinalities of frequent sets) is kept at its default of 3 and the confidence threshold for association rule mining for the Apriori algorithm~\cite{Agrawal_etal/1997} is lowered to 0.04 to increase options during rule selection.
To perform rule selection, CEGA filters the rules using another confidence threshold. 
To obtain discriminatory explanations, we used nine thresholds between $[0.05, \ldots, 0.95]$ and likewise, nine thresholds for characteristic explanations between $[0.05, \ldots, 0.8]$ (cf. Section~\ref{sec:relatedwork}).
Combined with the three different $\localexplainer$, we thus obtain 54 different rule models from CEGA for each black-box model it is tasked to explain. As with \algname, we select the rule model with the highest accuracy on $\instanceset$ and report results on the test data.

To compute local explanations, KS and LI were given a sampling budget of 300 and IG performed 200 steps to approximate the integral. 
The empirical mean of the black-box training data was used as baseline for all attribution methods.

\subsection{Algorithm Comparison: Properties (i)-(iv)}
\label{sec:comparison}
In this section, we compare {\algname} against {\CEGA} and the two variants of \anchors. First we show an example output of all three algorithms. 
Then, we present the results from four different aspects: faithfulness, comprehensibility, completeness, and robustness against the Rashomon effect (see Properties~(i)--(iv) in Section~\ref{sec:introduction}). 
Lastly, we compare the runtime of the rule extraction step for every algorithm. 

\begin{table}[!t]
\centering
\caption{Rules generated by {\algname}, {\anchors} and {\CEGA} for the same black-box model $\model$ on btsc. Labels \{0,1\} indicate whether an individual donated blood during a certain time period.}
\label{tab:rulesexample}

\resizebox{\columnwidth}{!}{
  \begin{tabular}{rl}
  \toprule
  \textbf{Algorithm} & \textbf{Rules} \\
  \midrule
  {\algname} & 
  $\begin{aligned}
  & \text{Precision} = 0.933,\ \text{F1} = 0.835\\
  0: & \ (\text{Amount} \in ( 250, 4\,250 ])\\
  1: & \ (\text{Recency} \in ( 1, 4 ]\land \text{Amount} \in ( 1\,250, 8\,500 ])
  \end{aligned}$ \\
  \midrule
  \anchors & 
  $\begin{aligned}
  & \text{Precision} = 0.893,\ \text{F1} = 0.893\\
  0: & \ (\text{Recency} \in ( 4.03, \infty]) \lor (\text{Amount} \in ( -\infty, 997.01 ]) \lor (\text{Time} \in ( 50.11, \infty]) \\
  1: & \ (\text{Amount} \in ( 997.01, \infty] \land \text{Recency} \in ( -\infty, 4.03 ] \land \text{Time} \in ( -\infty, 50.11 ]) \\
    & \ \lor (\text{Amount} \in ( 1\,930.35, \infty] \land \text{Recency} \in ( -\infty, 4.03 ])
  \end{aligned}$ \\
  \midrule
  {\CEGA} & 
  $\begin{aligned}
  & \text{Precision} = 0.176,\ \text{F1} = 0.036\\
  0: & \ (\text{Time} \in ( 14.82, 42.43 ]) \\
  1: & \ (\text{Frequency} \in ( 5.35, 8.86 ])
  \end{aligned}$ \\
  \bottomrule
  \end{tabular}
}
\vspace{-1.5em}
\end{table}

\textbf{Example Rule Models}
Table~\ref{tab:rulesexample} provides an example output on the btsc dataset for all three algorithms, each for the same black-box model.
The DNFs in the table are intended to illustrate compactness and comprehensibility. 
{\algname} and {\CEGA} contain 2 terms each (i.e., Size=2) with 2 or fewer literals each; {\anchors} provides 5 terms with up to 3 literals. {\anchors}' rules consist of half open intervals in this example. While half open intervals do not impede performance in terms of accuracy, we argue that it is questionable for an \textit{empirically obtained rule} to suggest that model behavior will not change before $\pm \infty$. Both {\algname} and {\CEGA}, by design, always provide bounded intervals.

Regarding the five properties listed in Section~\ref{sec:introduction}, consider Tables~\ref{tab:binarycomp} and \ref{tab:multicomp}.
They contain quantitative results on binary and multi-class tasks, respectively. 
For each dataset, the first three rows show the average values and the standard deviations of the F1-score, precisions, and rule sizes over the 50 black-box models. Recall that the rule size (Size) counts the number of terms across all classes for the final global rule model.
The fourth row (\xmark) denotes the proportion of the cases where the corresponding algorithm was able to produce rules  for every class with an overall better than random precision on the test set.
Since {\CEGA} is restricted to binary class problems, it is omitted from Table~\ref{tab:multicomp}.

\begin{table}[ht]

  \caption{Performance on binary tasks (\ref{tab:binarycomp}) and multi-class tasks (\ref{tab:multicomp}). 
  Size counts the number of terms in a global explanation across all classes. {\xmark} is the proportion of the 50 models for which
the algorithm provides a rule model with above chance precision and at least
one rule for each of the classes. 
All but the last row report mean$\pm$\footnotesize{std} over all 50 black-box models per task.
  {\anchorsinf} is the idealized version of {\anchors}.}

  \centering
  \begin{subtable}[t]{0.51\textwidth} 
    \centering
    \caption{Binary tasks}
    \resizebox{\linewidth}{!}{\begin{tabular}{@{}r@{}@{}c@{}*{3}{c}{c}@{}}
        \rotatebox{90}{\raisebox{\hb}{Task}} 
        & \raisebox{\mb}{Metric} & \raisebox{\mb}{\algname} & \raisebox{\mb}{\CEGA} & \raisebox{\mb}{\anchorsk} & \raisebox{\mb}{\anchorsinf} \\
\midrule
\multirow{7}{*}{\raisebox{\vb}{\rotatebox{90}{\raisebox{\hb}{btsc}}}} 
& F1 & 0.93\tiny{$\pm$0.02} & 0.09\tiny{$\pm$0.06} & 0.71\tiny{$\pm$0.26} & 0.90\tiny{$\pm$0.02}  \\
& Precision & 0.94\tiny{$\pm$0.02} & 0.54\tiny{$\pm$0.28} & 0.81\tiny{$\pm$0.15} & 0.90\tiny{$\pm$0.02} \\
& Size & 4.36\tiny{$\pm$1.54} & 2.22\tiny{$\pm$0.58} & 4.10\tiny{$\pm$0.99} & 5.58\tiny{$\pm$0.81}  \\
& \xmark & 0.98 & 0. & 0.3 & 0.7 \\
\midrule
\multirow{7}{*}{\raisebox{\vb}{\rotatebox{90}{\raisebox{\hb}{breastw}}}} 
& F1 & 0.96\tiny{$\pm$0.01} & 0.36\tiny{$\pm$0.06} & 0.79\tiny{$\pm$0.20} & 0.96\tiny{$\pm$0.02}  \\
& Precision & 0.97\tiny{$\pm$0.01} & 0.70\tiny{$\pm$0.03} & 0.90\tiny{$\pm$0.08} & 0.96\tiny{$\pm$0.02} \\
& Size & 15.12\tiny{$\pm$4.43} & 6.70\tiny{$\pm$2.84} & 13.76\tiny{$\pm$3.19} & 18.24\tiny{$\pm$3.51} \\
& \xmark & 1. & 0.02 & 0.72 & 1.  \\
\midrule
\multirow{7}{*}{\raisebox{\vb}{\rotatebox{90}{\raisebox{\hb}{heloc}}}} 
& F1 & 0.86\tiny{$\pm$0.02} & 0.20\tiny{$\pm$0.13}  & 0.56\tiny{$\pm$0.01} & 0.89\tiny{$\pm$0.01}\\
& Precision & 0.89\tiny{$\pm$0.01} & 0.76\tiny{$\pm$0.07} & 0.80\tiny{$\pm$0.06} & 0.89\tiny{$\pm$0.01} \\
& Size & 13.98\tiny{$\pm$10.55} & 3.98\tiny{$\pm$2.10} & 13.98\tiny{$\pm$10.55} & 68.48\tiny{$\pm$8.83} \\
& \xmark & 1. & 0.29  & 0. & 1.\\
\midrule
\multirow{7}{*}{\raisebox{\vb}{\rotatebox{90}{\raisebox{\hb}{spf}}}} 
& F1 & 0.76\tiny{$\pm$0.03} & 0.55\tiny{$\pm$0.05} & 0.27\tiny{$\pm$0.03} & 0.84\tiny{$\pm$0.02} \\
& Precision & 0.76\tiny{$\pm$0.03} & 0.57\tiny{$\pm$0.05} & 0.76\tiny{$\pm$0.05} & 0.85\tiny{$\pm$0.02} \\
& Size & 17.80\tiny{$\pm$6.01} & 43.34\tiny{$\pm$35.03} & 17.80\tiny{$\pm$6.01} & 70.10\tiny{$\pm$8.24} \\
& \xmark & 0.94 & 0.02 & 0. & 1. \\
\midrule
\multirow{7}{*}{\raisebox{\vb}{\rotatebox{90}{\raisebox{\hb}{breastc.}}}} 
& F1 & 0.88\tiny{$\pm$0.03} & 0.57\tiny{$\pm$0.07}  & 0.70\tiny{$\pm$0.05} & 0.93\tiny{$\pm$0.01}\\
& Precision & 0.92\tiny{$\pm$0.01} & 0.63\tiny{$\pm$0.02} & 0.98\tiny{$\pm$0.02} & 0.97\tiny{$\pm$0.01} \\
& Size & 3.34\tiny{$\pm$1.98} & 25.26\tiny{$\pm$10.71} & 3.34\tiny{$\pm$1.98} & 12.88\tiny{$\pm$1.77} \\
& \xmark & 1. & 0.1 & 0.06 & 1. \\
\midrule
\multirow{7}{*}{\raisebox{\vb}{\rotatebox{90}{\raisebox{\hb}{ionos.}}}} & F1 & 0.75\tiny{$\pm$0.05} & 0.53\tiny{$\pm$0.14} & 0.33\tiny{$\pm$0.09} & 0.81\tiny{$\pm$0.02} \\
& Precision & 0.78\tiny{$\pm$0.05} & 0.58\tiny{$\pm$0.10}  & 0.77\tiny{$\pm$0.11} & 0.85\tiny{$\pm$0.03}\\
& Size & 4.60\tiny{$\pm$1.44} & 67.32\tiny{$\pm$32.18}  & 4.60\tiny{$\pm$1.44} & 15.86\tiny{$\pm$2.86}\\
& \xmark & 0.86  & 0.04 & 0.08 & 1.\\

\midrule
\multirow{7}{*}{\raisebox{\vb}{\rotatebox{90}{\raisebox{\hb}{spambase}}}} 
& F1 & 0.87\tiny{$\pm$0.03} & 0.29\tiny{$\pm$0.15}  & 0.45\tiny{$\pm$0.10} & 0.76\tiny{$\pm$0.03}\\
& Precision & 0.90\tiny{$\pm$0.02} & 0.85\tiny{$\pm$0.06}  & 0.70\tiny{$\pm$0.07} & 0.76\tiny{$\pm$0.03}\\
& Size & 9.22\tiny{$\pm$6.95} & 8.42\tiny{$\pm$13.79}  & 8.66\tiny{$\pm$5.48} & 21.66\tiny{$\pm$2.13}\\
& \xmark & 1. & 0.14  & 0.1 & 1.\\
\bottomrule
        \end{tabular}}
    \label{tab:binarycomp}
  \end{subtable}
  \hspace{0.02\textwidth} 
  \begin{subtable}[t]{0.4\textwidth} 
    \centering
    \caption{Multi-class tasks}
    \resizebox{\linewidth}{!}{
                \begin{tabular}{@{}r@{}@{}c@{}*{3}{c}@{}}
        \rotatebox{90}{\raisebox{\hb}{Task}} & \raisebox{\mb}{Metric} & \raisebox{\mb}{\algname} & \raisebox{\mb}{\anchorsk} & \raisebox{\mb}{\anchorsinf}\\
        \midrule
\multirow{7}{*}{\raisebox{\vb}{\rotatebox{90}{\raisebox{\hb}{iris}}}} 
 & F1 & 0.87\tiny{$\pm$0.02} & 0.72\tiny{$\pm$0.08} & 0.92\tiny{$\pm$0.03} \\
 & Precision & 0.96\tiny{$\pm$0.01} & 0.94\tiny{$\pm$0.02} & 0.94\tiny{$\pm$0.02} \\
 & Size & 4.62\tiny{$\pm$0.73} & 4.62\tiny{$\pm$0.73} & 9.14\tiny{$\pm$0.95} \\
& \xmark & 1. & 0. & 1. \\
\midrule
\multirow{7}{*}{\raisebox{\vb}{\rotatebox{90}{\raisebox{\hb}{autouniv}}}} 
 & F1 & 0.41\tiny{$\pm$0.04} & 0.40\tiny{$\pm$0.06} & 0.54\tiny{$\pm$0.04} \\
 & Precision & 0.44\tiny{$\pm$0.04} & 0.65\tiny{$\pm$0.08} & 0.66\tiny{$\pm$0.05} \\
 & Size & 24.82\tiny{$\pm$5.60} & 24.80\tiny{$\pm$5.57} & 43.92\tiny{$\pm$4.63} \\
& \xmark & 0.96 & 0.36 & 1. \\
\midrule
\multirow{7}{*}{\raisebox{\vb}{\rotatebox{90}{\raisebox{\hb}{abalone}}}} 
 & F1 & 0.81\tiny{$\pm$0.02} & 0.77\tiny{$\pm$0.02} & 0.77\tiny{$\pm$0.02} \\
 & Precision & 0.86\tiny{$\pm$0.03} & 0.77\tiny{$\pm$0.02} & 0.77\tiny{$\pm$0.02} \\
 & Size & 25.24\tiny{$\pm$8.75} & 11.80\tiny{$\pm$2.81} & 11.80\tiny{$\pm$2.81} \\
& \xmark & 1. & 1. & 1. \\
\midrule
\multirow{7}{*}{\raisebox{\vb}{\rotatebox{90}{\raisebox{\hb}{wine}}}} 
 & F1 & 0.66\tiny{$\pm$0.08} & 0.50\tiny{$\pm$0.11} & 0.87\tiny{$\pm$0.04} \\
 & Precision & 0.85\tiny{$\pm$0.07} & 0.88\tiny{$\pm$0.07} & 0.95\tiny{$\pm$0.02} \\
 & Size & 4.30\tiny{$\pm$1.27} & 4.30\tiny{$\pm$1.27} & 10.44\tiny{$\pm$0.97} \\
& \xmark & 1. & 0.02 & 1. \\
\midrule
\multirow{7}{*}{\raisebox{\vb}{\rotatebox{90}{\raisebox{\hb}{vehicle}}}} 
 & F1 & 0.55\tiny{$\pm$0.04} & 0.57\tiny{$\pm$0.05} & 0.61\tiny{$\pm$0.03} \\
 & Precision & 0.61\tiny{$\pm$0.05} & 0.72\tiny{$\pm$0.04} & 0.70\tiny{$\pm$0.03} \\
 & Size & 19.48\tiny{$\pm$4.91} & 19.48\tiny{$\pm$4.91} & 36.60\tiny{$\pm$4.65} \\
& \xmark & 1. & 0.06 & 1. \\
\midrule
\multirow{7}{*}{\raisebox{\vb}{\rotatebox{90}{\raisebox{\hb}{beans}}}} 
 & F1 & 0.91\tiny{$\pm$0.02} & 0.80\tiny{$\pm$0.03} & 0.87\tiny{$\pm$0.01} \\
 & Precision & 0.93\tiny{$\pm$0.01} & 0.86\tiny{$\pm$0.02} & 0.87\tiny{$\pm$0.01} \\
 & Size & 34.50\tiny{$\pm$7.10} & 34.50\tiny{$\pm$7.10} & 63.98\tiny{$\pm$6.65} \\
& \xmark & 1. & 0.02 & 1. \\
\midrule
\multirow{7}{*}{\raisebox{\vb}{\rotatebox{90}{\raisebox{\hb}{diggle}}}} 
 & F1 & 0.85\tiny{$\pm$0.02} & 0.61\tiny{$\pm$0.05} & 0.71\tiny{$\pm$0.04} \\
 & Precision & 0.99\tiny{$\pm$0.01} & 0.71\tiny{$\pm$0.06} & 0.74\tiny{$\pm$0.04} \\
 & Size & 9.58\tiny{$\pm$0.64} & 9.58\tiny{$\pm$0.64} & 16.78\tiny{$\pm$1.76} \\
& \xmark & 1. & 0. & 0. \\

\bottomrule
        \end{tabular}
}
    \label{tab:multicomp}
  \end{subtable}
  \label{tab:algcomp}

\end{table}

\textbf{Properties (i) and (ii)} Regarding faithfulness and comprehensibility, a closer look at the F1-score and the precision results show that {\algname} always outperforms {\CEGA}, in most of the cases {\anchors}, and is comparable even to the idealized version of {\anchors} (i.e., {\anchorsinf}) in terms of faithfulness. In addition, {\algname}, almost always returns (much) more compact rules than {\anchorsinf} (see row Size). 

Comparing the idealized variant \anchorsinf\ to {\algname}, it outperforms \algname\ in terms of F1-score and precision in 8 out of 14 cases, while requiring up to $3 \times$ more terms at the same time. 
In 5 out of the 6 other datasets, {\algname} outperforms \anchorsinf\ on average, while requiring the same or only half the amount of terms. 
On one of the dataset (i.e., abalone), {\algname} needs more than twice the amount of terms (25.24 cmp. to 11.8). Comparing standard deviations, {\algname} and \anchorsinf\ actually overlap in half of the tasks.

Comparing against the normally budgeted \anchors, {\algname} outperforms it on average in 13 out of 14 cases, but with overlapping standard deviations in 5 cases. We note that both algorithms show notable gaps between F1-score and precision on the multi-class tasks. For {\anchors} this also holds true on the binary tasks, but not for {\algname}. In the results for {\CEGA}, the F1-scores and precision are notably lower compared to the other algorithms. It does provide the smallest rule models in 4 cases but also produces vastly larger rule models without notable performance gains in the other three cases.

\textbf{Property~(iii)}
In terms of completeness (row {\xmark}), {\algname} achieves a consistently high level of completeness, compared to the poor completeness results of {\CEGA} and {\anchorsk}. 
In particular, it only loses 11 models across three binary tasks and two in a single multi-class task out of the $2 \times 7 \times 50 = 700$ models.
{\anchorsk}, using the same size budget as {\algname}, loses models on all tasks except abalone, producing insufficient rule models in a total of 564 cases. Similarly, {\CEGA} fails to produce sufficient explanations in the vast majority of cases. When {\anchorsinf} is allowed to select as many rules as needed to reach maximum coverage, the number of insufficiently explained models drops to 65, though at the expense of increasing size by $2.2\times$ on average.

\textbf{Property~(iv)} Recall from Section~\ref{sec:prelimeval} that the robustness of a method against the Rashomon effect is measured by
the standard deviations of the mean F1-score and precision values.
The remarkably low standard deviations obtained for the F1-scores and precisions by {\algname} and {\anchorsinf} show that they are of comparably high stability with regard to the Rashomon effect. {\algname} matches or exceeds {\anchors} and {\CEGA} in all tasks.

More precisely, the standard deviation of the F1-scores is at most $0.05$ in most cases, it is $\leq$ 0.1 for {\CEGA} in three out of 7, and for {\anchorsk} in four out of 14 cases.
By providing similarly accurate rules for all black-boxes, both {\algname} and {\anchorsinf} show higher robustness against the Rashomon effect.

The standard deviation on the size indicates that all algorithms adapt to individual black-boxes by increasing complexity. For {\algname} and {\CEGA}, the standard deviation can exceed half the total value of the mean, whereas for {\anchorsinf}, it seldom exceeds 15\% of the mean.
This suggests a saturation of representation in {\anchorsinf}, where increasing the number of rules provides diminishing returns on novel information. In contrast, our top-down approach {\algname} adapts more effectively, providing new insights with fewer terms and avoiding the redundancy seen in bottom-up methods. 
\begin{figure*}[!t]
  
  \resizebox{\textwidth}{!}{\input{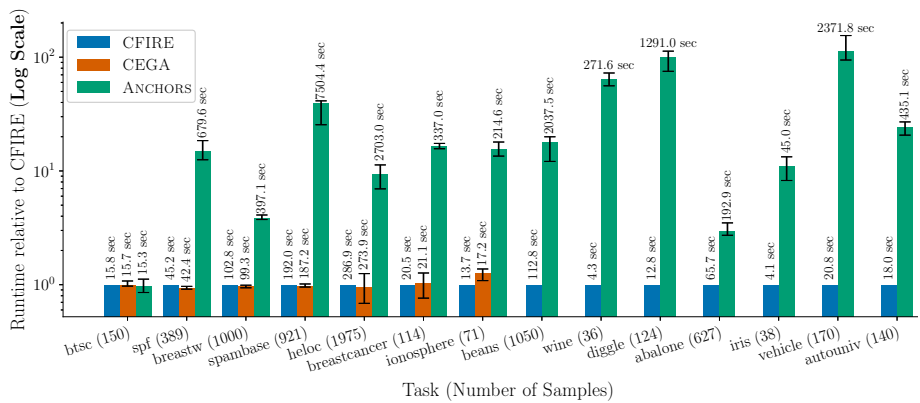}}
  \vspace{-2em}{
  \caption{Runtimes (Y-axis) of {\anchorsinf} and {\CEGA} \textit{relative} to {\algname} on the 14 datasets (X-axis). Note the \textit{log} scale of the Y-axis.
  The average runtime in seconds are provided atop the bars. Bar height is average runtime, whiskers indicate min-max values across the 50 black-boxes per task. 
  }
  }
  \label{fig:time}
  \vspace{-2em}
\end{figure*}

\textbf{Runtime} Figure~\hyperref[fig:time]{2} 
presents a runtime comparison for the initial rule extraction step. Each bar shows the average runtime of the respective algorithm \textit{relative} to \algname\
over all 50 models for each task on a \textit{logarithmic} scale. 
The black whiskers displayed atop each bar indicate minimum and maximum relative values, with absolute runtime in seconds above. 
\algname's and \CEGA's runtimes also account for the computation of all three local explainability methods. 
The runtime for \anchors\ is accumulated over all samples in $\instanceset$.
Rule selection is not contained in the runtime. Hence, we only report a single time for {\anchors} and \anchorsinf. \\
\indent The runtimes for {\algname} range from 4 to 290 seconds across all tasks. Where applicable, CEGA demonstrates comparable performance. 
\anchors\ is faster only on btsc than {\algname} for certain models, taking on average 0.5 seconds less. 
For all other datasets, {\algname} is faster by a factor of $2.95$ to $113$.
For all algorithms, the runtime depends not only on the cardinality of $\instanceset$, but also on the dimensionality of the data. For example, consider the runtimes for breastw $(d=9)$ and spambase $(d=57)$. 
For these tasks, where $|\instanceset| \approx 1\,000$, \algname\ takes twice as long on spambase (192 sec) compared to breastw (102 sec). 
While this increase is notable, \anchors\ takes 397 seconds on average on breastw, but 7\,504 seconds (i.e., more than two hours) on spambase. 
For a $6.3\times$ increase in dimensionality this reflects an $18.9\times$ increase in runtime. Overall, both \algname\ and \CEGA\ offer compelling runtime advantages over \anchors.

\textbf{Summary} The results show a significant runtime advantage of the top-down methods compared to \anchors, despite them calculating multiple global explanations based on different local explainability methods. {\CEGA} falls behind in the faithfulness measures and shows mixed results regarding compactness.
Our algorithm \algname\ is well equipped to provide explanations in binary as well as multi-class settings, providing high precision rules and insights into all classes in nearly all cases. The results indicate that our approach is \textit{robust} against the Rashomon effect,
effectively adapting the number of terms used to describe a black-box model. {\anchorsinf}, the theoretically unbounded version of {\anchors}, shows similar stability. However, as expected, the increase in complexity is disproportionately higher.

\subsection{Flexibility: Property (v)}
\label{sec:disagreement}

\begin{figure}[!t]
  \centering
  \resizebox{\textwidth}{!}{\input{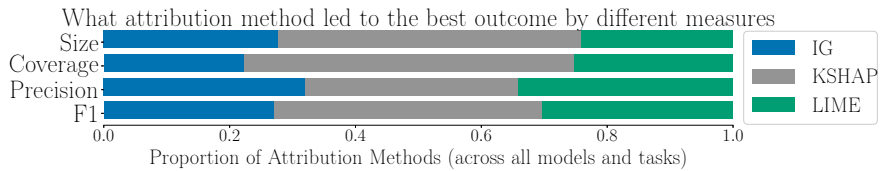}}
  \vspace{-2em}{
  \label{fig:disagreement}
  \caption{Summarized over all tasks and black-boxes, the \textit{frequency} of each attribution method leading to the best model for a given measure (Size, Coverage, Precision or F1).}}
  
\vspace{-2em}
\end{figure}

In the analysis above we computed rule models for all three attribution methods and chose the best. 
We now move on to analyze the impact of flexibility (Property~(v)) on the results for {\algname} by asking the following questions: Which $\localexplainer \in \{ \text{KS, LI, IG}\}$ would have been the best choice, according to target measures other than accuracy on the test set? How large would the performance difference be if we used a single method?
For that, Figure~\hyperref[fig:disagreement]{3} 
provides a summarizing overview for F1-score, precision, coverage, and size as target measures. The plot aggregates over all datasets. For this plot we filter out results for all black-box models that were not explained using any $\localexplainer$. 
This completely removes explanations for 4 of the 700 black-boxes we trained. Recall that {\algname} lost 13 models, meaning that for 13-4=9 models it would have provided an explanation but they were not chosen by our selection criterion of highest accuracy. Of the 696 remaining black-boxes, only 120 were best explained by a single attribution method according to all measures simultaneously. In those 120 cases KS led to the best outcome in 52\% of cases, IG in 27\% and LI in 21\%. While KS represents the majority of the cases, in 48\% of cases a different method would have been better overall.
Looking at the individual measures in Figure~\hyperref[fig:disagreement]{3} across all models, KS dominates in terms of F1-score, coverage, and size, but the number of cases where LI and IG are the best choice is non-negligible. For precision, all methods are equally represented. This already demonstrates the beneficiality of flexibility.

\textbf{Example Rule Models}
Table~\ref{tab:disagexample} shows examples of rule models computed by {\algname} given IG and LI as $\localexplainer$ for the same black-box model. One can see that the different local explainers lead to semantically different rule models. More precisely, for the binary task btsc, we see that both methods lead to rule models with Size=2. For LI, the rules rely on the features `Amount' and `Recency', while for IG, the rules also use `Time'. Most notably, `Recency' is used strictly in different classes by both rule models. They achieve very similar F1-scores.
The examples for the iris dataset are also quite different. Rule models for LI and IG use 4 and 5 terms, each across all three classes. For class Setosa, IG leads to three terms with a single literal each, focusing on different intervals on the Sepal Width feature. In comparison, LI produces a single term that evaluates both Petal Length and Width.

\begin{table}[!t]
\centering
\caption{{\algname} Rules generated using IG or LI for the same black-boxes on btsc and iris.}
\label{tab:disagexample}

\resizebox{\columnwidth}{!}{
  \begin{tabular}{ll}
  \toprule
  
  \textbf{btsc} & \\
  LI & Precision = 0.911, F1 = 0.899 \\
  & $\begin{aligned}
  
  \text{0}: & \ (\text{Amount} \in ( 250, 4\,250 ]) \\
  \text{1}: & \ (\text{Recency} \in ( 1, 4 ] \land \text{Amount} \in ( 1250, 8\,500 ])
  \end{aligned}$ \\ \\
  IG & Precision = 0.904, F1 = 0.895 \\
  & $\begin{aligned}
  \text{0}: & \ (\text{Recency} \in ( 0, 23 ] \land \text{Time} \in ( 2, 98 ]) \\
  \text{1}: & \ (\text{Amount} \in ( 4\,625, 8\,500 ])
  \end{aligned}$ \\
  
  \midrule
  \textbf{iris} & \\
  LI & Precision = 0.939, F1 = 0.886 \\
  & $\begin{aligned}
  \text{Setosa}: & \ (\text{Petal Length} \in ( 1, 1.7 ] \land \text{Petal Width} \in ( 0.1, 0.5 ]) \\
  \text{Versicolour}: & \ (\text{Sepal Length} \in ( 5.85, 6.6 ]) \ \lor (\text{Sepal Width} \in ( 2.2, 2.95 ]) \\
  \text{Virginica}: & \ (\text{Petal Length} \in ( 4.8, 6.7 ] \land \text{Petal Width} \in ( 1.6, 2.4 ])
  \end{aligned}$ \\ \\
  
  IG & Precision = 0.933, F1 = 0.835 \\
  & $\begin{aligned}
  \text{Setosa}: & \ (\text{Sepal Width} \in ( 3.05, 3.15 ]) \lor (\text{Sepal Width} \in ( 3.35, 3.7 ]) \lor (\text{Sepal Width} \in ( 3.85, 4.2 ]) \\
  \text{Versicolour}: & \ (\text{Sepal Length} \in ( 5.20, 6.65 ]) \\
  \text{Virginica}: & \ (\text{Petal Length} \in ( 4.80, 6.7 ] \land \text{Petal Width} \in ( 1.6, 2.4 ])
  \end{aligned}$ \\
  
  \bottomrule
  \end{tabular}
}

\vspace{-2em}
\end{table}

\begin{table}[ht]
  \caption{Comparison of individual attribution methods against {\algname} on binary and multi-class tasks. Size and {\xmark} are the same as in Table~\ref{tab:algcomp}. 
  Prec. $\localexplainer$  quantifies the agreement
between rule models and local explanations.
  All rows but {\xmark} report mean$\pm$\footnotesize{std} over all 50 black-box models per task.}
  \centering
  \begin{subtable}[t]{0.46\textwidth} 
    \centering
    \caption{Binary tasks}
    \resizebox{\linewidth}{!}{ \begin{tabular}{@{}r@{}@{}c@{}*{4}{c}@{}}

        \rotatebox{90}{\raisebox{\hb}{Task}} & \raisebox{\mb}{Metric} & \raisebox{\mb}{\algname} & \raisebox{\mb}{\algname-KS} & \raisebox{\mb}{\algname-LI} & \raisebox{\mb}{\algname-IG}\\
        \midrule
        \multirow{7}{*}{\raisebox{\vb}{\rotatebox{90}{\raisebox{\hb}{\centering btsc}}}} 
 & F1 & 0.93\tiny{$\pm$0.02} & 0.92\tiny{$\pm$0.02} & 0.91\tiny{$\pm$0.02} & 0.90\tiny{$\pm$0.01} \\
 & Precision & 0.94\tiny{$\pm$0.02} & 0.93\tiny{$\pm$0.02} & 0.93\tiny{$\pm$0.02} & 0.91\tiny{$\pm$0.01} \\
 & Size & 4.36\tiny{$\pm$1.54} & 3.78\tiny{$\pm$1.15} & 3.56\tiny{$\pm$1.80} & 2.42\tiny{$\pm$1.31} \\
        & \xmark & 0.98 & 0.98 & 0.98 & 0.92 \\
 & Prec. $\localexplainer$& 0.72\tiny{$\pm$ 0.09} & 0.77\tiny{$\pm$ 0.02} & 0.58\tiny{$\pm$ 0.05} & 0.98\tiny{$\pm$ 0.06} \\
\midrule
\multirow{7}{*}{\raisebox{\vb}{\rotatebox{90}{\raisebox{\hb}{spf}}}} 
 & F1 & 0.76\tiny{$\pm$0.03} & 0.74\tiny{$\pm$0.03} & 0.73\tiny{$\pm$0.03} & 0.76\tiny{$\pm$0.03} \\
 & Precision & 0.76\tiny{$\pm$0.03} & 0.74\tiny{$\pm$0.03} & 0.73\tiny{$\pm$0.03} & 0.77\tiny{$\pm$0.03} \\
 & Size & 17.80\tiny{$\pm$6.01} & 18.02\tiny{$\pm$5.69} & 14.34\tiny{$\pm$4.76} & 13.76\tiny{$\pm$5.48} \\
& \xmark & 0.94 & 0.56 & 0.56 & 0.94 \\
 & Prec. $\localexplainer$ & 0.78\tiny{$\pm$ 0.14} & 0.72\tiny{$\pm$ 0.09} & 0.51\tiny{$\pm$ 0.10} & 0.83\tiny{$\pm$ 0.12} \\
\midrule
\multirow{7}{*}{\raisebox{\vb}{\rotatebox{90}{\raisebox{\hb}{breastw}}}} 
 & F1 & 0.96\tiny{$\pm$0.01} & 0.97\tiny{$\pm$0.01} & 0.95\tiny{$\pm$0.02} & 0.91\tiny{$\pm$0.03} \\
 & Precision & 0.97\tiny{$\pm$0.01} & 0.97\tiny{$\pm$0.01} & 0.96\tiny{$\pm$0.02} & 0.94\tiny{$\pm$0.03} \\
 & Size & 15.12\tiny{$\pm$4.43} & 15.68\tiny{$\pm$4.27} & 10.02\tiny{$\pm$3.41} & 2.94\tiny{$\pm$1.02} \\
& \xmark & 1. & 1. & 1. & 0.76 \\
 & Prec. $\localexplainer$& 0.74\tiny{$\pm$ 0.22} & 0.87\tiny{$\pm$ 0.02} & 0.35\tiny{$\pm$ 0.04} & 0.49\tiny{$\pm$ 0.29} \\
\midrule
\multirow{7}{*}{\raisebox{\vb}{\rotatebox{90}{\raisebox{\hb}{spam}}}} 
 & F1 & 0.87\tiny{$\pm$0.03} & 0.83\tiny{$\pm$0.07} & 0.83\tiny{$\pm$0.01} & 0.87\tiny{$\pm$0.01} \\
 & Precision & 0.90\tiny{$\pm$0.02} & 0.83\tiny{$\pm$0.07} & 0.87\tiny{$\pm$0.04} & 0.88\tiny{$\pm$0.01} \\
 & Size & 9.22\tiny{$\pm$6.95} & 16.06\tiny{$\pm$7.48} & 8.84\tiny{$\pm$3.15} & 4.90\tiny{$\pm$1.30} \\
& \xmark & 1. & 0.64 & 1. & 1. \\
 & Prec. $\localexplainer$& 0.87\tiny{$\pm$ 0.21} & 0.73\tiny{$\pm$ 0.07} & 0.47\tiny{$\pm$ 0.05} & 0.99\tiny{$\pm$ 0.00} \\
\midrule
\multirow{7}{*}{\raisebox{\vb}{\rotatebox{90}{\raisebox{\hb}{heloc}}}} 
 & F1 & 0.86\tiny{$\pm$0.02} & 0.87\tiny{$\pm$0.03} & 0.87\tiny{$\pm$0.01} & 0.85\tiny{$\pm$0.02} \\
 & Precision & 0.89\tiny{$\pm$0.01} & 0.87\tiny{$\pm$0.03} & 0.88\tiny{$\pm$0.01} & 0.88\tiny{$\pm$0.02} \\
 & Size & 13.98\tiny{$\pm$10.55} & 23.28\tiny{$\pm$7.40} & 8.40\tiny{$\pm$5.29} & 10.52\tiny{$\pm$5.58} \\
& \xmark & 1. & 1. & 1. & 1. \\
 & Prec. $\localexplainer$& 0.85\tiny{$\pm$ 0.15} & 0.71\tiny{$\pm$ 0.04} & 0.57\tiny{$\pm$ 0.08} & 0.92\tiny{$\pm$ 0.06} \\
\midrule
\multirow{7}{*}{\raisebox{\vb}{\rotatebox{90}{\raisebox{\hb}{breastc.}}}} 
 & F1 & 0.88\tiny{$\pm$0.03} & 0.89\tiny{$\pm$0.04} & 0.82\tiny{$\pm$0.07} & 0.88\tiny{$\pm$0.03} \\
 & Precision & 0.92\tiny{$\pm$0.01} & 0.90\tiny{$\pm$0.04} & 0.85\tiny{$\pm$0.06} & 0.92\tiny{$\pm$0.02} \\
 & Size & 3.34\tiny{$\pm$1.98} & 6.96\tiny{$\pm$2.05} & 5.58\tiny{$\pm$2.17} & 2.66\tiny{$\pm$1.08} \\
& \xmark & 1. & 1. & 1. & 1. \\
 & Prec. $\localexplainer$& 0.96\tiny{$\pm$ 0.13} & 0.88\tiny{$\pm$ 0.04} & 0.31\tiny{$\pm$ 0.07} & 0.98\tiny{$\pm$ 0.04} \\
\midrule
\multirow{7}{*}{\raisebox{\vb}{\rotatebox{90}{\raisebox{\hb}{\centering iono.}}}} 
 & F1 & 0.75\tiny{$\pm$0.05} & 0.74\tiny{$\pm$0.06} & 0.86\tiny{$\pm$0.03} & 0.74\tiny{$\pm$0.05} \\
 & Precision & 0.78\tiny{$\pm$0.05} & 0.74\tiny{$\pm$0.06} & 0.87\tiny{$\pm$0.02} & 0.78\tiny{$\pm$0.05} \\
 & Size & 4.60\tiny{$\pm$1.44} & 2.46\tiny{$\pm$1.01} & 4.36\tiny{$\pm$0.96} & 4.56\tiny{$\pm$1.45} \\
& \xmark & 0.86 & 0.02 & 0.58 & 0.88 \\
 & Prec. $\localexplainer$& 0.96\tiny{$\pm$ 0.07} & 0.93\tiny{$\pm$ 0.07} & 0.63\tiny{$\pm$ 0.18} & 0.97\tiny{$\pm$ 0.02} \\
\bottomrule
        \end{tabular}}
    \label{tab:disagbinary}
  \end{subtable}
  \hspace{0.02\textwidth}
  \begin{subtable}[t]{0.45\textwidth}
    \centering
    \caption{Multi-class tasks}
    \resizebox{\linewidth}{!}{\begin{tabular}{@{}r@{}@{}c@{}*{4}{c}@{}}
        \rotatebox{90}{\raisebox{\hb}{Task}} & \raisebox{\mb}{Metric} & \raisebox{\mb}{\algname} & \raisebox{\mb}{\algname-KS} & \raisebox{\mb}{\algname-LI} & \raisebox{\mb}{\algname-IG}\\
        \midrule
        \multirow{7}{*}{\raisebox{\vb}{\rotatebox{90}{\raisebox{\hb}{beans}}}} 
 & F1 & 0.91\tiny{$\pm$0.02} & 0.92\tiny{$\pm$0.01} & 0.89\tiny{$\pm$0.03} & 0.91\tiny{$\pm$0.01} \\
 & Precision & 0.93\tiny{$\pm$0.01} & 0.93\tiny{$\pm$0.01} & 0.92\tiny{$\pm$0.02} & 0.93\tiny{$\pm$0.01} \\
 & Size & 34.50\tiny{$\pm$7.10} & 36.68\tiny{$\pm$6.16} & 36.64\tiny{$\pm$6.50} & 27.32\tiny{$\pm$4.23} \\
        & \xmark & 1. & 1. & 1. & 1. \\
 & Prec. $\localexplainer$& 0.88\tiny{$\pm$0.06} & 0.89\tiny{$\pm$0.02} & 0.82\tiny{$\pm$0.04} & 0.93\tiny{$\pm$0.03} \\
\midrule
\multirow{7}{*}{\raisebox{\vb}{\rotatebox{90}{\raisebox{\hb}{wine}}}} 
 & F1 & 0.66\tiny{$\pm$0.08} & 0.66\tiny{$\pm$0.07} & 0.70\tiny{$\pm$0.07} & 0.40\tiny{$\pm$0.04} \\
 & Precision & 0.85\tiny{$\pm$0.07} & 0.86\tiny{$\pm$0.06} & 0.85\tiny{$\pm$0.05} & 0.53\tiny{$\pm$0.06} \\
 & Size & 4.30\tiny{$\pm$1.27} & 3.96\tiny{$\pm$1.03} & 5.08\tiny{$\pm$1.26} & 5.12\tiny{$\pm$1.33} \\
& \xmark & 1. & 1. & 1. & 1. \\
 & Prec. $\localexplainer$& 0.96\tiny{$\pm$0.03} & 0.96\tiny{$\pm$0.03} & 0.93\tiny{$\pm$0.04} & 0.98\tiny{$\pm$0.03} \\
\midrule
\multirow{7}{*}{\raisebox{\vb}{\rotatebox{90}{\raisebox{\hb}{diggle}}}} 
 & F1 & 0.85\tiny{$\pm$0.02} & 0.85\tiny{$\pm$0.02} & 0.84\tiny{$\pm$0.02} & 0.68\tiny{$\pm$0.06} \\
 & Precision & 0.99\tiny{$\pm$0.01} & 0.99\tiny{$\pm$0.01} & 0.98\tiny{$\pm$0.04} & 0.88\tiny{$\pm$0.07} \\
 & Size & 9.58\tiny{$\pm$0.64} & 9.58\tiny{$\pm$0.64} & 10.42\tiny{$\pm$0.70} & 12.94\tiny{$\pm$1.46} \\
& \xmark & 1. & 1. & 0.96 & 1. \\
 & Prec. $\localexplainer$& 0.99\tiny{$\pm$0.01} & 0.99\tiny{$\pm$0.01} & 0.90\tiny{$\pm$0.05} & 0.96\tiny{$\pm$0.03} \\
\midrule
\multirow{7}{*}{\raisebox{\vb}{\rotatebox{90}{\raisebox{\hb}{abalone}}}} 
 & F1 & 0.81\tiny{$\pm$0.02} & 0.81\tiny{$\pm$0.02} & 0.80\tiny{$\pm$0.03} & 0.75\tiny{$\pm$0.02} \\
 & Precision & 0.86\tiny{$\pm$0.03} & 0.83\tiny{$\pm$0.02} & 0.85\tiny{$\pm$0.04} & 0.77\tiny{$\pm$0.03} \\
 & Size & 25.24\tiny{$\pm$8.75} & 31.62\tiny{$\pm$6.26} & 21.84\tiny{$\pm$5.79} & 19.00\tiny{$\pm$5.57} \\
& \xmark & 1. & 1. & 1. & 1. \\
 & Prec. $\localexplainer$& 0.76\tiny{$\pm$0.07} & 0.71\tiny{$\pm$0.07} & 0.77\tiny{$\pm$0.08} & 0.90\tiny{$\pm$0.04} \\
\midrule
\multirow{7}{*}{\raisebox{\vb}{\rotatebox{90}{\raisebox{\hb}{iris}}}} 
 & F1 & 0.87\tiny{$\pm$0.02} & 0.87\tiny{$\pm$0.02} & 0.87\tiny{$\pm$0.03} & 0.83\tiny{$\pm$0.01} \\
 & Precision & 0.96\tiny{$\pm$0.01} & 0.96\tiny{$\pm$0.01} & 0.96\tiny{$\pm$0.02} & 0.93\tiny{$\pm$0.02} \\
 & Size & 4.62\tiny{$\pm$0.73} & 4.62\tiny{$\pm$0.73} & 3.06\tiny{$\pm$0.24} & 5.92\tiny{$\pm$0.27} \\
& \xmark & 1. & 1. & 1. & 1. \\
 & Prec. $\localexplainer$& 0.94\tiny{$\pm$0.09} & 0.94\tiny{$\pm$0.09} & 0.92\tiny{$\pm$0.09} & 0.94\tiny{$\pm$0.03} \\
\midrule
\multirow{7}{*}{\raisebox{\vb}{\rotatebox{90}{\raisebox{\hb}{vehicle}}}}  & F1 & 0.55\tiny{$\pm$0.04} & 0.53\tiny{$\pm$0.04} & 0.52\tiny{$\pm$0.04} & 0.56\tiny{$\pm$0.04} \\
 & Precision & 0.61\tiny{$\pm$0.05} & 0.56\tiny{$\pm$0.05} & 0.58\tiny{$\pm$0.05} & 0.61\tiny{$\pm$0.04} \\
 & Size & 19.48\tiny{$\pm$4.91} & 22.70\tiny{$\pm$5.27} & 21.06\tiny{$\pm$3.83} & 17.42\tiny{$\pm$3.92} \\
& \xmark & 1. & 0.94 & 0.98 & 1. \\

 & Prec. $\localexplainer$& 0.91\tiny{$\pm$0.07} & 0.74\tiny{$\pm$0.08} & 0.80\tiny{$\pm$0.08} & 0.93\tiny{$\pm$0.03} \\
\midrule
\multirow{7}{*}{\raisebox{\vb}{\rotatebox{90}{\raisebox{\hb}{autouniv}}}} 
 & F1 & 0.41\tiny{$\pm$0.04} & 0.43\tiny{$\pm$0.04} & 0.40\tiny{$\pm$0.04} & 0.43\tiny{$\pm$0.04} \\
 & Precision & 0.44\tiny{$\pm$0.04} & 0.44\tiny{$\pm$0.05} & 0.43\tiny{$\pm$0.04} & 0.46\tiny{$\pm$0.04} \\
 & Size & 24.82\tiny{$\pm$5.60} & 23.02\tiny{$\pm$5.28} & 24.50\tiny{$\pm$5.35} & 19.80\tiny{$\pm$4.27} \\
& \xmark & 0.96 & 0.96 & 0.94 & 0.98 \\
 & Prec. $\localexplainer$& 0.70\tiny{$\pm$0.08} & 0.73\tiny{$\pm$0.07} & 0.65\tiny{$\pm$0.07} & 0.75\tiny{$\pm$0.08}\\
\bottomrule
        \end{tabular}}
    \label{table1b}
    \label{tab:disagmulti}
  \end{subtable}
  
\vspace{-2em}
\end{table}

\textbf{Faithfulness and Compactness} In addition to the qualitative example above, Tables \ref{tab:disagbinary} and \ref{tab:disagmulti} quantitatively confirm that there is indeed a notable downstream effect caused by the disagreement problem. Column~{\algname} is given as reference, again depicting average scores for the best result according to accuracy on $\instanceset$. Columns {\algname}-\{KS, LI, IG\} show results for the individual feature attribution methods. 

The last three columns
show that restricting {\algname} to a \textit{single} local explainer may result in entirely different outputs. 
Notable performance differences with respect to F1-score and precision are observed on 8 datasets. Explanations based on {\algname}-IG underperform notably on breastw, wine, diggle, abalone, and iris, with performance differences of up to 25\% compared to {\algname}. {\algname}-LI deviates negatively on spambase and breastcancer, but actually outperforms {\algname} on ionosphere. On ionosphere, both {\algname}-KS and {\algname}-IG perform equally but worse than {\algname}-LI. Overall, {\algname}-KS delivers solid accuracy and is seldom outperformed by a large margin.\\
\indent For {\algname}-KS and {\algname}-LI, Size is very similar in 9 cases. In the other cases {\algname}-LI provides an explanation with less terms than {\algname}-KS.
However, IG leads to the most compact explanation in nearly all cases. Most importantly, it does so in cases where it does not underperform regarding the other measures. 

\textbf{Completeness} 
Across all 14 tasks, KS leads to a loss (\xmark) of 95 (out of 700) models, LI 50, and IG 26 compared to just 12 for {\algname}.
Conspicuously, the majority of insufficiently explained models does not stem from the multi-class tasks but in fact from the binary tasks. 
For spf, breastw, spambase, and ionosphere, at least one attribution method fails on 10 or more models.

\textbf{Robustness against the Rashomon Effect} 
Regarding the standard deviation across models, we note that it remains small for F1-score and precision. It exceeds 5\% in only 5 cases. The standard deviation for Size is equal or lower for individual $\localexplainer$ than for \algname, but still exceeds variability of \anchorsinf, reaching 20\% or more of the total mean value even for datasets with a comparably large size such as spf, beans or abalone.
This clearly demonstrates that part of \algname's higher robustness against the Rashomon effect is due to its flexibility to work with different local explainers.

\textbf{(Dis)agreement between Rules and Local Explanations (Prec. $\localexplainer$)} Recall that this property is measured by the precision with which the applicable rules match the dimensions that are important according to the local explanation, provided that the rules are consistent with the black-box model on the sample.
Overall we observe a large agreement between the local explanations and the final rules but there are differences between the individual local explainers. For {\algname}-KS and {\algname}-IG, the precision is always above 0.7 and often even above 0.8, whereas for {\algname}-LI it only exceeds 0.7 in 6 cases and at the lowest drops down to 0.31. {\algname} reaches Prec. {$\localexplainer$} of $\geq$ 0.85 on seven datasets, with values > 0.7 in the other cases. {\algname}, naturally, seldom reaches the top score for each row, however, it surprisingly ranks at least second in all but four cases. The overall good alignment on test data of nearly all methods indicates that the rules successfully capture key semantic properties of the local explanations. This indicates that a high train accuracy (our selection criterion) is related to a good alignment, also validating {\algname}'s bias towards high-coverage boxes during the greedy rule selection step.

\textbf{Summary} Comparing {\algname} variants using a single local explainer we found: {\algname}-IG lost the least amount of models but also showed the least reliable performance regarding accuracy. 
{\algname}-KS provides strong results regarding accuracy in all cases but struggled with {\xmark} the most. 
{\algname}-LI underperformed in accuracy on 3 datasets only, but it did loose double the amount of models compared to {\algname}-IG.

Regarding {\xmark}, each local explainer failed notably on at least one task. 
While all methods can be summarised by distinctive strengths and weaknesses, \xmark\ indicates that, regardless what other metric might be regarded most important in a certain context, the Rashomon effect is only accounted for if results from multiple local explainers are considered. 
This implies that the flexibility to choose different local explainers is a key property for robustness of local-to-global explainers.

\newpage
\section{Concluding Remarks}
\label{sec:conclusion}

We presented {\algname}, a novel XAI algorithm computing global rule models from local explanations for black-box models in form of a set of class DNFs.
We experimentally demonstrated that, in contrast to the state-of-the-art methods, it fulfills all five desirable properties outlined in Section~\ref{sec:introduction}.
In particular, \algname\ computed faithful, compact, and complete DNFs for nearly all black-box models.
The flexibility to choose among different local explainers proved to be a key property to {\algname}'s success in several cases.
It was as fast as {\CEGA} and up to two magnitudes faster than {\anchors}.
In comparison, {\anchors} struggled with faithfulness and completeness, {\anchorsinf}, the idealized version of {\anchors}, required considerably more rules, and {\CEGA} failed in every regard.
{\algname} exhibited the greatest adaptability in adjusting its rule complexity to different black-box models, demonstrating its ability to adjust to model-specific characteristics when needed.
In assessing {\algname}'s flexibility, our experiments, to the best of our knowledge, also provide the first sizable empirical evidence of the downstream effects of the disagreement problem. Additionally, our evaluation scheme builds on recent results on the Rashomon effect in XAI and our results underscore that accounting for this effect is crucial for an adequat experimental evaluation of explanation methods.

While {\algname} provided good precision scores in nearly all cases, its coverage could be improved in multi-class settings. One possible solution  could be to allow {\algname} to generate synthetic samples and query the black-box to enrich the data available, similar to how {\anchors} also explores the data space locally. 

Although the greedy set cover heuristic proved effective at the rule composition stage, it would be interesting to see how closed itemset mining and more sophisticated composition approaches like GLocalX~\cite{setzu2021glocalx} interplay.

Our ablation study on the downstream effects of the disagreement problem showed that each local explainer that we considered has led to distinct outputs. We picked a particular local explainer based on a rule model's predictive performance on the instance set, but as we also saw, a single local explainer seldom led to the best outcome according to other criteria. Further research is needed to understand the specific qualitative differences in attribution methods that make one particularly effective for generating strong explanations with desired properties for specific tasks or models. Alternatively, future work could explore combining explanations from different methods before itemset mining, rather than running {\algname} separately for each explainer.

The rules produced by {\algname} for different classes can overlap (see Table~\ref{tab:rulesexample} for an example).
In this work, we resolve ambiguity for specific instances by assigning the class of the empirically more accurate rule (cf. Section~\ref{sec:algorithm}). 
While our experimental results justify this approach, the trade-off between faithfulness (avoiding ambiguity) and comprehensibility (remaining compact) should be explored further.
One way could be to associate the rules with a reliability or certainty score for conflicting areas.
While not resolving the fundamental conflict, it provides additional information to the user. 
We are working on a more sophisticated approach based on van Fraassen's model for explanations~\cite{VanFraassen/1980}.

\small{
\subsubsection*{Acknowledgments}
This research has been funded by the Federal Ministry of Education and Research of Germany and the state of North-Rhine Westphalia as part of the Lamarr-Institute for Machine Learning and Artificial Intelligence Lamarr22B.
}
\vspace*{-1.5em}

\bibliographystyle{splncs04}

\begin{thebibliography}{10}
\providecommand{\url}[1]{\texttt{#1}}
\providecommand{\urlprefix}{URL }
\providecommand{\doi}[1]{https://doi.org/#1}

\bibitem{dry_bean_602}
{Dry Bean}. UCI Machine Learning Repository (2020), \url{https://doi.org/10.24432/C50S4B}

\bibitem{wine_109}
Aeberhard, S., Forina, M.: {Wine}. UCI Machine Learning Repository (1992), \url{https://doi.org/10.24432/C5PC7J}

\bibitem{agarwal2022openxai}
Agarwal, C., Krishna, S., Saxena, E., Pawelczyk, M., Johnson, N., Puri, I., Zitnik, M., Lakkaraju, H.: {Openxai: Towards a transparent evaluation of model explanations}. Advances in neural information processing systems  \textbf{35},  15784--15799 (2022)

\bibitem{aggarwal2014frequent}
Aggarwal, C.C., Han, J.: Frequent Pattern Mining. Springer Cham, 1 edn. (2014)

\bibitem{Agrawal_etal/1997}
Agrawal, R., Mannila, H., Srikant, R., Toivonen, H., Verkamo, A.I.: Fast discovery of association rules. In: Fayyad, U.M., Piatetsky{-}Shapiro, G., Smyth, P., Uthurusamy, R. (eds.) Advances in Knowledge Discovery and Data Mining, pp. 307--328. {AAAI/MIT} Press (1996)

\bibitem{alkhatib2022cega}
Alkhatib, A., Bostr{\"o}m, H., Vazirgiannis, M.: {Explaining predictions by characteristic rules}. In: Joint European Conference on Machine Learning and Knowledge Discovery in Databases. pp. 389--403. Springer (2022)

\bibitem{beckh2022hackfleisch}
Beckh, K., M{\"u}ller, S., R{\"u}ping, S.: {A Quantitative Human-Grounded Evaluation Process for Explainable Machine Learning.} In: LWDA. pp. 13--20 (2022)

\bibitem{boley2010listing}
Boley, M., Horv{\'{a}}th, T., Poign{\'{e}}, A., Wrobel, S.: Listing closed sets of strongly accessible set systems with applications to data mining. Theor. Comput. Sci.  \textbf{411}(3),  691--700 (2010)

\bibitem{breiman2001statistical}
Breiman, L.: {Statistical modeling: The two cultures (with comments and a rejoinder by the author)}. Statistical science  \textbf{16}(3),  199--231 (2001)

\bibitem{steel_plates_faults_198_spf}
Buscema, M, T.S., Tastle, W.: {Steel Plates Faults}. UCI Machine Learning Repository (2010), \url{https://doi.org/10.24432/C5J88N}

\bibitem{cormen2022introduction}
Cormen, T.H., Leiserson, C.E., Rivest, R.L., Stein, C.: {Introduction to algorithms}. MIT press, 4 edn. (2022)

\bibitem{diggle1990time}
Diggle, P.J.: {Time series: a biostatistical introduction}. Oxford University Press (1990)

\bibitem{iris_53}
Fisher, R.A.: {Iris}. UCI Machine Learning Repository (1936), \url{https://doi.org/10.24432/C56C76}

\bibitem{gely2005generic}
G{\'e}ly, A.: {A generic algorithm for generating closed sets of a binary relation}. In: Formal Concept Analysis: Third International Conference, ICFCA 2005, Lens, France, February 14-18, 2005. Proceedings 3. pp. 223--234. Springer (2005)

\bibitem{guidotti2019lore}
Guidotti, R., Monreale, A., Giannotti, F., Pedreschi, D., Ruggieri, S., Turini, F.: {Factual and counterfactual explanations for black box decision making}. IEEE Intelligent Systems  \textbf{34}(6),  14--23 (2019)

\bibitem{autouniv}
Hickey, R.: {AutoUniv}. Online database (2010), \url{https://www.openml.org/d/1553}, accessed: 2024-09-19

\bibitem{spambase_94}
Hopkins, Mark, R.E.F.G., Suermondt, J.: {Spambase}. UCI Machine Learning Repository (1999), \url{https://doi.org/10.24432/C53G6X}

\bibitem{jacovi2020towards}
Jacovi, A., Goldberg, Y.: {Towards Faithfully Interpretable NLP Systems: How Should We Define and Evaluate Faithfulness?} In: Proceedings of the 58th Annual Meeting of the Association for Computational Linguistics. pp. 4198--4205 (2020)

\bibitem{breastw}
{Jan van Rijn}: {BNG (breast-w) Dataset}. Online database (2014), \url{https://www.openml.org/d/251}, accessed: 2024-09-19

\bibitem{krishna2022disagreement}
Krishna, S., Han, T., Gu, A., Pombra, J., Jabbari, S., Wu, S., Lakkaraju, H.: {The disagreement problem in explainable machine learning: A practitioner's perspective}. arXiv preprint arXiv:2202.01602  (2022)

\bibitem{lakkaraju2016decisionsets}
Lakkaraju, H., Bach, S.H., Leskovec, J.: {Interpretable decision sets: A joint framework for description and prediction}. In: Proceedings of the 22nd ACM SIGKDD international conference on knowledge discovery and data mining. pp. 1675--1684 (2016)

\bibitem{letham2015stroke}
Letham, B., Rudin, C., McCormick, T.H., Madigan, D.: {Interpretable classifiers using rules and bayesian analysis: Building a better stroke prediction model}  (2015)

\bibitem{lundberg2017SHAP}
Lundberg, S.M., Lee, S.I.: {A unified approach to interpreting model predictions}. Advances in neural information processing systems  \textbf{30} (2017)

\bibitem{molnar2022interpretableml}
Molnar, C.: {Interpretable Machine Learning}. 2 edn. (2022), \url{https://christophm.github.io/interpretable-ml-book}

\bibitem{statlog_(vehicle_silhouettes)_149}
Mowforth, P., Shepherd, B.: {Statlog (Vehicle Silhouettes)}. UCI Machine Learning Repository, \url{https://doi.org/10.24432/C5HG6N}

\bibitem{mueller2023rashomon}
M{\"u}ller, S., Toborek, V., Beckh, K., Jakobs, M., Bauckhage, C., Welke, P.: {An empirical evaluation of the Rashomon effect in explainable machine learning}. In: Joint European Conference on Machine Learning and Knowledge Discovery in Databases. pp. 462--478. Springer (2023)

\bibitem{abalone_1}
Nash, Warwick, S.T.T.S.C.A., Ford, W.: {Abalone}. UCI Machine Learning Repository (1994), \url{https://www.openml.org/d/183}, {DOI}: https://doi.org/10.24432/C55C7W

\bibitem{nauta2023co12}
Nauta, M., Trienes, J., Pathak, S., Nguyen, E., Peters, M., Schmitt, Y., Schl{\"o}tterer, J., Van~Keulen, M., Seifert, C.: {From anecdotal evidence to quantitative evaluation methods: A systematic review on evaluating explainable ai}. ACM Computing Surveys  \textbf{55}(13s),  1--42 (2023)

\bibitem{pasquier1999discovering}
Pasquier, N., Bastide, Y., Taouil, R., Lakhal, L.: Efficient mining of association rules using closed itemset lattices. Inf. Syst.  \textbf{24}(1),  25--46 (1999)

\bibitem{pitt1988computational}
Pitt, L., Valiant, L.G.: Computational limitations on learning from examples. J. {ACM}  \textbf{35}(4),  965--984 (1988)

\bibitem{ribeiro2016LIME}
Ribeiro, M.T., Singh, S., Guestrin, C.: {``Why should i trust you?" Explaining the predictions of any classifier}. In: Proceedings of the 22nd ACM SIGKDD international conference on knowledge discovery and data mining. pp. 1135--1144 (2016)

\bibitem{ribeiro2018anchors}
Ribeiro, M.T., Singh, S., Guestrin, C.: {Anchors: High-precision model-agnostic explanations}. In: Proceedings of the AAAI conference on artificial intelligence. vol.~32 (2018)

\bibitem{setzu2021glocalx}
Setzu, M., Guidotti, R., Monreale, A., Turini, F., Pedreschi, D., Giannotti, F.: {Glocalx-from local to global explanations of black box ai models}. Artificial Intelligence  \textbf{294},  103457 (2021)

\bibitem{ionosphere_52}
Sigillito, V., W.S.H.L., Baker, K.: {Ionosphere}. UCI Machine Learning Repository (1989), \url{https://doi.org/10.24432/C5W01B}

\bibitem{sundararajan2017axiomatic}
Sundararajan, M., Taly, A., Yan, Q.: {Axiomatic attribution for deep networks}. In: International conference on machine learning. pp. 3319--3328. PMLR (2017)

\bibitem{VanFraassen/1980}
Van~Fraassen, B.C.: The Scientific Image. Oxford University Press, New York (1980)

\bibitem{wang2015falling}
Wang, F., Rudin, C.: {Falling rule lists}. In: Artificial intelligence and statistics. pp. 1013--1022. PMLR (2015)

\bibitem{breast_cancer_wisconsin_(diagnostic)_17}
Wolberg, William, M.O.S.N., Street, W.: {Breast Cancer Wisconsin (Diagnostic)}. UCI Machine Learning Repository (1993), \url{https://doi.org/10.24432/C5DW2B}

\bibitem{btsc_176}
Yeh, I.C.: {Blood Transfusion Service Center}. UCI Machine Learning Repository (2008), \url{https://doi.org/10.24432/C5GS39}

\end{thebibliography}

\end{document}